\pdfoutput=1

\documentclass[11pt]{article}

\usepackage[preprint]{acl}
\usepackage[utf8]{inputenc}

\usepackage{booktabs}  
\usepackage{array}     
\usepackage{times}
\usepackage{latexsym}
\usepackage{tabularx}
\newcolumntype{Y}{>{\centering\arraybackslash}X}
\usepackage{array,tabularx}
\newcolumntype{L}{>{\raggedright\arraybackslash}X}
\usepackage{svg}
\usepackage{amsmath}
\usepackage{cancel}
\usepackage{ulem}
\usepackage{bm}
\usepackage{pgfplots}
\usepackage{pgfplotstable}
\pgfplotsset{compat=newest}
\usepackage[english]{babel}
\usetikzlibrary{calc}
\usepackage{xcolor}
\usepackage{amssymb}
\usepackage{pdfpages}

\usepackage[inline]{enumitem}

\graphicspath{{assets/}}

\usepackage[T1]{fontenc}

\usepackage{microtype}

\usepackage{inconsolata}

\usepackage{graphicx}

\title{Reading Between the Signs:\\Predicting Future Suicidal Ideation from Adolescent Social Media Texts}

\author{
  Paul Blum\textsuperscript{\textdagger},
  Enrico Liscio\textsuperscript{\textdagger}, 
  Ruixuan Zhang\textsuperscript{\textdagger},\\
  \textbf{Caroline Figueroa\textsuperscript{\textdagger}\textsuperscript{\textdaggerdbl},
  and Pradeep K. Murukannaiah\textsuperscript{\textdagger}}
  \\
  \textsuperscript{\textdagger}Delft University of Technology \hspace{1cm}
  \textsuperscript{\textdaggerdbl}Stanford University \\
  \texttt{paul-blum@outlook.com}
  \\
  \texttt{\{e.liscio, r.zhang-2, c.figueroa, p.k.murukannaiah\}@tudelft.nl}
}

\begin{document}

\maketitle
\begin{abstract}
Suicide is a leading cause of death, yet predicting it remains a significant challenge. Risk factors such as depression or substance use are commonly used for prediction, but their predictive performance is often only slightly better than chance. Additionally, many cases go undetected due to a lack of contact with mental health services. Social media, however, offers a unique opportunity, as people often share their thoughts and struggles online in real time. In this work, we propose a novel task and method for early identification: predicting suicidal ideation and behavior (SIB) \textit{before} a user ever expresses it on an online forum. 
We introduce \textsc{Early-SIB}, a transformer-based model that sequentially processes the posts a user writes and engages with to predict whether they will write a SIB post. Our model achieves a balanced accuracy of 0.73 in predicting future SIB on a Dutch youth forum, demonstrating that such tools can offer a meaningful addition to traditional methods. Finally, we use Shapley Additive Explanations to make the model's predictions more interpretable.
\end{abstract}

\section{Introduction}
Suicide is a pressing issue, globally being the third leading cause of death among 15--29-year-olds \cite{world_health_organization_suicide_2024}. Moreover, each suicide is estimated to meaningfully affect 135 individuals around the victim \cite{cerel_how_2019}. However, predicting suicide is a significant public health challenge. 60\% of people who die by suicide have not expressed suicidal ideation as documented in their medical records or inquired through a semi-structured interview \cite{mchugh_association_2019}. Furthermore, our ability to predict suicide has not improved over the past 50 years \cite{franklin_risk_2017}. 
Thus, there is an urgent need for innovative methods to detect suicide risk in adolescents. 

Research on suicide has mainly focused on assessing a person's risk of suicide using so-called ``risk factors'' \cite{turecki_suicide_2019}. These are traits or conditions thought to increase the likelihood of suicidal behavior. These factors have informed widely adopted guidelines \cite{american_association_of_suicidology_know_2023,world_health_organization_suicide_2024,american_foundation_for_suicide_prevention_risk_2025, national_institute_of_mental_health_nimh_warning_2025}. For adolescents, risk factors associated with suicide include sex, depression, anxiety, adverse childhood experiences, self-hate, and self-injury \cite{cheng_relevant_2025}. However, a meta-analysis of decades of suicide research reveals that risk factor-based prediction performs only slightly better than chance \cite{franklin_risk_2017}. The guidelines often lack specificity, classifying nearly anyone with mental illness, chronic physical illness, life stress, special population status, or access to lethal means as being at risk \cite{franklin_risk_2017}. 

Suicidal Ideation and Behaviour (SIB) refers to thoughts, plans, or actions related to suicide \cite{posner_columbia-suicide_2010}. Ideation is one of the earliest and typically unobserved stages in the progression of an individual's suicidal process \cite{ wasserman_european_2012}. SIB, which also includes suicide attempts, is one of the strongest predictors of death by suicide \cite{klonsky_suicide_2016}. If we can predict SIB early, we could enable more timely intervention, which we view as a positive societal impact insofar as it increases the likelihood that at-risk individuals can be identified and offered support at an earlier stage, potentially saving lives.

Given the widespread use of social media and increased self-disclosure during the years of adolescence \cite{vijayakumar_self-disclosure_2020}, social media data could prove to be a valuable source for identifying suicide risk and SIB in teenagers. Teenagers often do not disclose risk factors to their physicians; instead, many share them on social media \cite{pourmand_social_2019}, often making peers the first recipients of their distress calls \cite{belfort_similarities_2012, vijayakumar_self-disclosure_2020}. Additionally, over two-thirds of people who die by suicide have had no contact with professional mental health care in the year before their death \cite{stene-larsen_contact_2019}, essentially making them overlooked by traditional healthcare systems. 
Even for those who seek professional help, the time between healthcare interactions is immense compared to social media interactions \cite{coppersmith_natural_2018}. 

We address a novel challenge to approach suicide prediction in adolescents: \textbf{predicting future SIB}---i.e., whether an adolescent will exhibit SIB---based solely on their social media post history.
To this end, we employ a supervised Machine Learning (ML) approach, which has the potential to provide superior predictive accuracy compared to theory-driven approaches  \cite{info:doi/10.2196/73052}.
However, unlike existing ML approaches, our approach does not rely on any self-disclosure of SIB as input to the model but instead treats it as the label to predict, effectively moving the prediction one step earlier in the timeline. 

We introduce \textbf{\textsc{Early-SIB}}, a transformer-based model designed to predict future SIB from the posts that a user has written and interacted with. We experiment with \textsc{Early-SIB} on De Kindertelefoon\footnote{\url{https://forum.kindertelefoon.nl/} \label{kt-fn}}, a Dutch adolescent help forum, and show that SIB can be predicted from a user's post history.
We then employ SHAP, an explainable AI technique, to reveal how the model consistently relies on a combination of the posts to make its predictions, rather than on a few significant posts, underscoring the complexity of the task.
These findings suggest that ML tools applied to social media posts may complement traditional risk factor-based approaches for suicide prediction.

\begin{figure*}[ht]
  \centering
  \includegraphics[width=\textwidth]{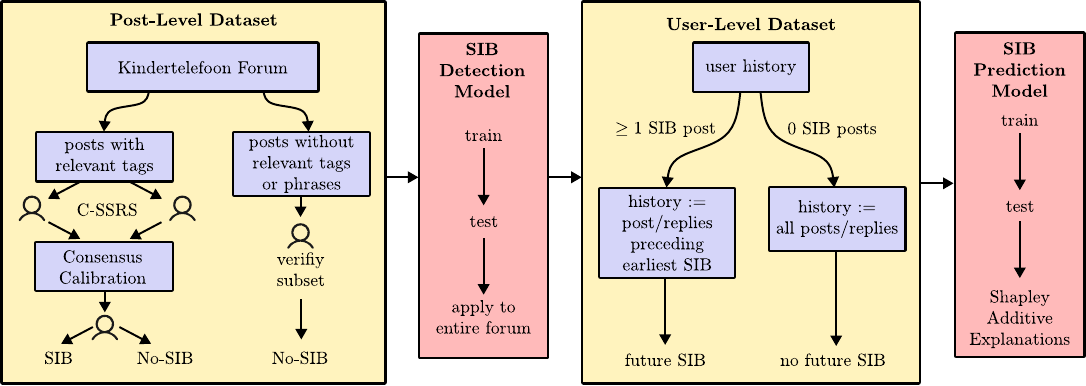}
  \caption{Pipeline of our approach. We manually label a small post-level dataset, which we use to train a model to detect SIB. We apply this model to the entire forum to get a user-level dataset. Finally, we train a model to predict future SIB and use SHAP to explain the predictions.}
  \label{fig:pipeline}
\end{figure*}

\section{Related Works}
We first review traditional clinical methods of identifying suicide risk before turning to recent work using natural language processing (NLP). 

\subsection{Clinical Methods}
\label{sec:related-works:traditional-methods}
The \citet{us_food_and_drug_administration_guidance_2012} considers the Columbia Suicide Severity Rating Scale (C-SSRS) to be the gold standard for suicide risk assessment in clinical trials. 
C-SSRS provides a list of questions, e.g., ``have you had any thoughts about killing yourself?'', to identify suicidal ideation and behavior \cite{posner_columbia-suicide_2010}. 
The European Psychiatric Association \cite{wasserman_european_2012} recommends that suicide risk be assessed through clinical interviews that evaluate the patient's psychological and social functioning. This includes assessments of personality, evaluations of socio-economic status, and the use of psychometric instruments such as scales \cite{hamilton_rating_1960, paykel_life_1976, linehan_reasons_1983,patterson_evaluation_1983,posner_columbia-suicide_2010}. There are also biological measurements, but they are only possible at specialized clinics and mainly used in research settings \cite{wasserman_european_2012}.


\subsection{NLP for Suicide Detection}
Recent years have seen growing interest in using NLP to assess suicide risk. Many studies use patients' electronic medical records 
\cite{carson_identification_2019,bittar_text_2019,zhong_use_2019,tsui_natural_2021,singh_rawat_parameter_2022,zandbiglari_enhancing_2025} or chat messages between help-seeks and providers \cite{bialer_detecting_2022, broadbent_machine_2023,qiu_psyguard_2024}. However, these approaches neglect people who never seek help or have limited contact with healthcare professionals.

Another approach is using social media platforms like Twitter and Reddit, e.g., by classifying posts into containing suicidal thoughts or not \cite{cao_latent_2019, mathur_utilizing_2020,roy_machine_2020,haque_comparative_2022} or classifying users into different risk levels of suicide \cite{mohammadi_clac_2019,sawhney_time-aware_2020, sawhney_phase_2021, sawhney_towards_2021}. Datasets have supported this direction by providing Reddit posts which have been expert-labeled into risk categories \cite{zirikly_clpsych_2019,gaur_knowledge-aware_2019, park_suicidal_2020}. While these approaches demonstrate the feasibility of detecting users at risk, they do not predict future disclosure of SIB. This limits their utility for early prevention or broader screening. 

A smaller number of studies have attempted to infer suicide risk early. \citet{coppersmith_natural_2018} and work in the CLPsych Task 2021 \cite{macavaney_community-level_2021, gollapalli_suicide_2021, gamoran_using_2021, morales_team_2021, wang_learning_2021} used social media posts from users with actual suicide attempts. However, such data 
may already include explicit self-reports of suicide,  limiting its usefulness for early detection. In contrast, we aim to predict self-reports before the user ever shares them, moving one step earlier in the timeline.

Recent efforts under the CLPsych Task 2024 \cite{chim_overview_2024} have been made to provide explanations of the model's risk prediction. They 
highlight salient sub-phrases within suicidal Reddit posts and summarize that evidence at the user level \cite{koushik_detecting_2024, alhamed_using_2024, gyanendro_singh_extracting_2024}. In contrast, our work explains predictions in an early prediction setting, where we highlight earlier interactions that do not contain SIB but are indicative of future risk.

\subsection{NLP for Mental Health Detection}
\label{sec:related-works:mental-health}
Aside from SIB, NLP has been used for early detection of mental health issues, including depression \cite{zogan_explainable_2022,wang_esdm_2024,hadzic_enhancing_2024}, eating disorders \cite{lopez-ubeda_how_2021, marmol_romero_mentalriskes_2024}, and gambling addiction \cite{perrot_development_2022,smith_automatic_2024}, with one of the most important forums being ERisk@CLEF \cite{montejo-raez_survey_2024}.

We draw on architectural ideas from \citet{zhang_natural_2024}, who proposed a model for detecting users with depression on social media. While they use posts directly indicative of depression as input, their architecture proves to be a valuable starting point for our task, because it is capable of taking multiple posts from a user's history as input.

\begin{figure*}[ht]
  \centering
  \includegraphics[width=\textwidth]{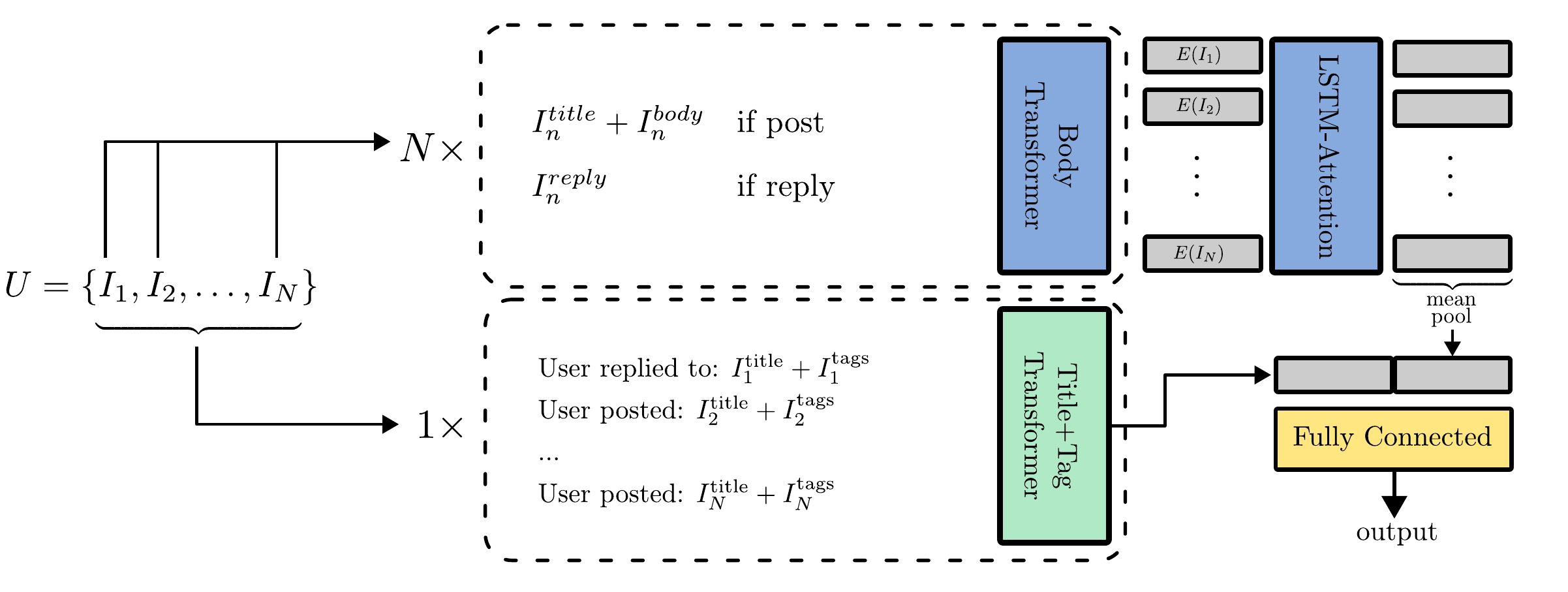}
  \caption{Architecture of our proposed model \textsc{Early-SIB} for early prediction of suicidal ideation and behaviour.}
  \label{fig:architecture}
\end{figure*}

\section{Dataset}
\label{sec:data}

We use data from the Kindertelefoon\textsuperscript{\ref{kt-fn}}, a Dutch forum explicitly dedicated as a space where adolescents (12--18 years of age) can seek support and share concerns related to mental health and wellbeing. Unlike broader platforms such as Reddit, where users often divide their activity across unrelated subforums, the Kindertelefoon provides a concentrated environment in which discussions remain focused on these issues. This focus enables frequent, meaningful, and in-depth conversations, making the dataset a particularly valuable resource for research on mental health struggles \cite{kuber-etal-2025-signs}.

We first signed a collaboration agreement with the Kindertelefoon, including permission to scrape their forum (which is publicly available and users are anonymous), and obtained approval from the ethics committee of the home university of the leading author.
We then extracted over 37,000 forum posts, along with all corresponding replies. All usernames were pseudonimized. We proceeded by labeling a small \textit{post-level} dataset, which we use to train a model to detect SIB posts. This model is then used to create a \textit{user-level} dataset that excludes any SIB posts. Data and annotations will be available under restricted access, as per agreement with the Kindertelefoon.

\subsection{Post-Level Dataset}
\label{sec:post-level-dataset}
Figure~\ref{fig:pipeline} (left) shows the labeling process for the post-level dataset. This dataset is intended to train a model to detect whether a post contains SIB or not, as described in Section~\ref{sec:experimental:sub:identifying}.
Specifically, we retrieved posts in which users had included any of the following tags: `suicide', `suicidal thoughts', `thinking of suicide', `suicidal', or `taking my life'. These posts were then annotated following the consensus calibration process \cite{oortwijn_interrater_2021} as follows. 
\begin{enumerate*}[label=(\arabic*)]
    \item Two annotators familiarized themselves with the annotation guidelines (the C-SRSS guidelines introduced in Section~\ref{sec:related-works:traditional-methods}).
    \item The key concepts of suicidal ideation and suicidal behavior were discussed. 
    \item The two annotators individually labeled the same randomly selected subset of 100 posts. 
    \item The annotations were compared, resulting in a Cohen's $\kappa = 0.66$ (substantial agreement). 
    \item The annotators met to discuss the posts where they disagreed. After the disagreement resolution, perfect agreement was reached ($\kappa = 1.00$).
    \item The rest of the annotations were performed by one of the annotators, resulting in a total of 569 posts labeled as SIB and 138 labeled as No-SIB. We refer to these 138 posts as ``hard'' instances, as they contain language relating to suicide but are labeled as No-SIB.
\end{enumerate*} 
See Appendix~\ref{app:annotation} for more details on the annotation procedure.

To generate more instances of No-SIB, we randomly sampled an additional 1,300 posts from the entire forum, ensuring that these posts did not contain any of the previously mentioned tags in their title, body, or tags, nor did they include the spans `I can't anymore', `dead', `end to', or `kill'. A validation of 200 such posts revealed that only one contained SIB. The final post-level dataset contained 569 SIB (28\%) and 1438 No-SIB (72\%) posts.

\subsection{User-Level Dataset}
We use the post-level dataset to train a classifier to detect SIB posts (see Section~\ref{sec:experimental:sub:identifying}). 
The classifier is applied to the entire forum data, assigning a SIB/No-SIB label to each post. From this, we create a user-level dataset, containing for each user a label indicating whether they have shared at least one SIB post or not, as well as that user's interaction history up to but excluding any SIB post (Figure~\ref{fig:pipeline}, right). This results in 284 users labeled SIB and 7,256 users labeled No-SIB.
For our early prediction task, the input is exclusively the interactions that came \textbf{before} any SIB post, and the goal is to predict whether a given user will go on to share SIB or not.

\section{Methodology}
\label{methodology}

We introduce \textsc{Early-SIB}, a model for predicting SIB early (Figure~\ref{fig:architecture}).
With each data point in the user-level dataset being a user's entire history, 50\% of inputs exceed the maximum length allowed by BERT-based models, and 25\% surpass the limit for LLaMA models.
To be able to process a user's entire history, we draw architectural inspiration from \citet{zhang_natural_2024} (see Section~\ref{sec:related-works:mental-health}). 

\textsc{Early-SIB} takes as input a user $U$ represented by $N$ forum interactions $I_1 \dots I_N$ (where an interaction is either a post or a reply). Each interaction (title+body in case of a post and the reply text in case of a reply) is then passed into a pretrained BERT model (the Body Transformer). 
We obtain $N$ CLS representations, which are passed into a bidirectional LSTM followed by an attention layer. This enables the model to capture sequential dependencies among interactions while assigning greater weight to the most informative ones. We mean pool across the output vectors, and this is fed as one part into a fully connected layer.
Additionally, the titles and tags that a user interacts with carry important clues for the prediction, which is why we concatenate all titles and tags into one string. This is fed into another pretrained BERT model (the Title+Tag Transformer), and its CLS embedding forms the other part fed into the fully connected layer. We use BERTje \cite{vries_bertje_2019}, a Dutch BERT model, to initialize the model weights.
\section{Experimental Setup}
\label{sec:experimental}
Our experiments consist of three parts. First, we curate a user-level dataset by training a model to \textit{detect} SIB in forum posts (Section~\ref{sec:experimental:sub:identifying}). Second, we train a model to \textit{predict} whether a user will share SIB based on their forum interactions that precede their SIB post (Section~\ref{sec:experimental:sub:predicting}). Third, we leverage an explainable AI technique to identify the most relevant post from the user’s history that contributed to the prediction (Section~\ref{sec:experimental:sub:explainability}).  
Appendix~\ref{appendix:experiments} provides additional experimental details, including hyperparameters and computational resources.
The code is provided as supplemental material. 

\subsection{Detecting SIB Posts}
\label{sec:experimental:sub:identifying}
We compare prompting-based and supervised approaches to SIB detection, which consists of a binary classification of SIB presence in a forum post. 
Prompting methods include zero-shot inference on LLaMA-3-8B-Instruct \cite{touvron_llama_2023} using either a simple prompt briefly describing the task or a prompt including the C-SSRS guideline (see Appendix~\ref{app:prompts}). We test both English and Dutch versions of the prompt. We also instruction-tune the models with the same prompts. The supervised methods include finetuning XLM-RoBERTa \cite{conneau_unsupervised_2020} and LLaMA-3-8B for binary sequence classification.
We report the results as the weighted $F_1$-score over 5-fold cross-validation on the post-level dataset. We use the best-performing method to assign SIB/No-SIB labels to all forum posts to generate the user-level dataset.

\subsection{Predicting Future SIB}
\label{sec:experimental:sub:predicting}
We evaluate \textsc{Early-SIB} on the user-level dataset.
We compare it to two baselines: zero-shot prompting on LLaMA-3-8B-Instruct and the architecture presented by \citet{zhang_natural_2024}.  We report results over a 5-fold cross-validation.
We limit the number of interactions used as input to our model to $N = 30$, prioritizing posts over replies in filling the 30 slots. $N=30$ covers more than 90\% of users in our data. We also experiment with different $N$'s.

The dataset is heavily imbalanced, with less than 4\% of users belonging to the SIB class and having at least one interaction prior to a SIB post. We found that down-sampling the majority class in the training set to a 50-50 distribution yielded the best performance, compared to using weighted loss, and different factors for down-sampling the majority class or up-sampling the minority class (see Appendix~\ref{app:hyper:prediction} for all tested resampling ratios).

\subsection{Explaining the Prediction}
\label{sec:experimental:sub:explainability}
Given the sensitivity of the task, it is crucial to provide transparency in predicting future SIB. We use Shapley additive explanations (SHAP, \citet{lundberg_unified_2017}) to interpret the model's predictions. SHAP is a game-theoretic approach that assigns an importance value to each feature (in our case, each user interaction) based on its contribution to the final output. We use it to identify common patterns (e.g., recency) in the impact that different interactions have when predicting future SIB.

\section{Results and Discussion}
\label{sec:results_and_discussion}

We report the results of SIB detection and future SIB prediction, with an analysis of the \textsc{Early-SIB} method through ablation studies and SHAP.

\subsection{Detecting SIB Posts}
\begin{figure}[t]
    \centering
    \begin{minipage}[b]{0.38\columnwidth}
        \centering
        \begin{tikzpicture}
            \begin{axis}[
                title={\small\textbf{Detection}},
                colormap={bluewhite}{color=(white) rgb255=(90,96,191)},
                xlabel=Predicted,
                ylabel=Actual,
                xtick={0,1},
                ytick={0,1},
                xticklabels={No SIB, SIB},
                yticklabels={No SIB, SIB},
                xticklabel style={rotate=0, anchor=north, font=\small},
                yticklabel style={rotate=90, anchor=south, font=\small},
                xtick style={draw=none},
                ytick style={draw=none},
                enlargelimits=false,
                point meta max=1.00,
                point meta min=0.00,
                width=1.8cm,
                height=1.8cm,
                scale only axis=true,
                axis on top,
                every axis label/.append style={font=\small},
                tick label style={font=\small},
                axis equal,
            ]
            \addplot[
                matrix plot,
                mesh/cols=2,
                point meta=explicit,
                draw=gray
            ] table [meta=C] {
                x y C
                0 0 0.97
                1 0 0.03
                0 1 0.09
                1 1 0.91
            };
            \node at (axis cs:0,0) {\small\textcolor{white}{$0.97$}};
            \node at (axis cs:1,0) {\small\textcolor{black}{$0.03$}};
            \node at (axis cs:0,1) {\small\textcolor{black}{$0.09$}};
            \node at (axis cs:1,1) {\small\textcolor{white}{$0.91$}};
            \end{axis}
        \end{tikzpicture}
    \end{minipage}
    \begin{minipage}[b]{0.56\columnwidth}
        \centering
        \begin{tikzpicture}
            \begin{axis}[
                title={\small\textbf{Early Prediction}},
                title style={yshift=-0.3ex}, 
                colormap={bluewhite}{color=(white) rgb255=(90,96,191)},
                xlabel=Predicted,
                ylabel=Actual,
                xtick={0,1},
                ytick={0,1},
                xticklabels={No SIB,SIB},
                yticklabels={No SIB,SIB},
                xticklabel style={rotate=0, anchor=north, font=\small},
                yticklabel style={rotate=90, anchor=south, font=\small},
                xtick style={draw=none},
                ytick style={draw=none},
                enlargelimits=false,
                colorbar,
                colorbar style={
                    width=0.2cm,
                    yticklabel style={font=\small}
                },
                point meta max=1.00,
                point meta min=0.00,
                width=1.8cm,
                height=1.8cm,
                scale only axis=true,
                axis on top,
                every axis label/.append style={font=\small},
                tick label style={font=\small},
                axis equal,
            ]
            \addplot[
                matrix plot,
                mesh/cols=2,
                point meta=explicit,
                draw=gray
            ] table [meta=C] {
                x y C
                0 0 0.74
                1 0 0.26
                0 1 0.29
                1 1 0.71
            };
            \node at (axis cs:0,0) {\small\textcolor{white}{$0.74$}};
            \node at (axis cs:1,0) {\small\textcolor{black}{$0.26$}};
            \node at (axis cs:0,1) {\small\textcolor{black}{$0.29$}};
            \node at (axis cs:1,1) {\small\textcolor{white}{$0.71$}};
            \end{axis}
        \end{tikzpicture}
    \end{minipage}
    \caption{Confusion matrices for best performance on detecting SIB (left) and early prediction of SIB (right). Row-normalized and averaged over evaluation folds.}
    \label{fig:confusion-matrices}
\end{figure}
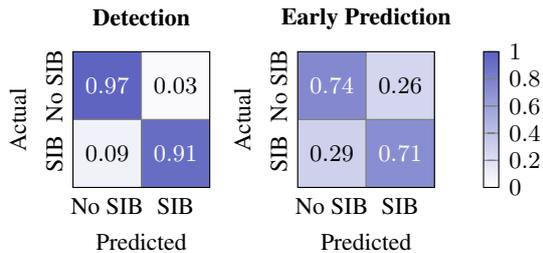

Table~\ref{tab:detecting_sib} summarizes performances in detecting SIB. The best-performing method is finetuning LLaMA-3-8B for binary sequence classification ($F_1$-score = $0.96\pm0.01$, see Figure~\ref{fig:confusion-matrices} (left) for the corresponding confusion matrix). LLaMA-3-8B also yields an $F_1$-score of $0.88\pm0.02$ on the ``hard'' subset of test instances where the inputs contain language around suicide but are not actual self-reports of SIB (see Section~\ref{sec:post-level-dataset}). This confirms the model’s ability to pick up and distinguish language related to suicide.

\begin{table}[!h]
\small
\begin{tabularx}\linewidth{@{}lYYY@{}}
\toprule
\textbf{Method} & \textbf{$F_1$-score} & \textbf{Recall} & \textbf{Precision} \\
\midrule
L3-ZS-EN-S & $0.74 \pm 0.02$ & $0.34 \pm 0.02$ & $0.71 \pm 0.03$ \\
L3-ZS-EN-C & $0.74 \pm 0.02$ & $0.31 \pm 0.03$ & $0.77 \pm 0.04$ \\
L3-ZS-NL-S & $0.61 \pm 0.02$ & $0.02 \pm 0.01$ & $0.87 \pm 0.16$ \\
L3-ZS-NL-C & $0.60 \pm 0.02$ & $0.02 \pm 0.01$ & $0.32 \pm 0.17$ \\
L3-IT-EN-S & $0.93 \pm 0.01$ & $0.88 \pm 0.05$ & $0.86 \pm 0.04$ \\
L3-IT-EN-C & $0.92 \pm 0.02$ & $0.89 \pm 0.06$ & $0.85 \pm 0.05$ \\
L3-IT-NL-S & $0.92 \pm 0.01$ & $0.84 \pm 0.05$ & $0.89 \pm 0.02$ \\
L3-IT-NL-C & $0.73 \pm 0.15$ & $0.36 \pm 0.43$ & $0.62 \pm 0.18$ \\
XLM-roberta & $0.91 \pm 0.02$ & $0.89 \pm 0.02$ & $0.81 \pm 0.06$ \\
\textbf{L3-SC} & $0.96 \pm 0.02$ & $0.91 \pm 0.04$ & $0.92\pm0.01$ \\
\bottomrule
\end{tabularx}

\caption{SIB detection performance. Abbreviations: L3 = LLaMA-3, ZS = Zero-Shot, IT = Instruction-Tuned, SC = Sequence Classification Head, EN = English, NL = Dutch, S = Simple Prompt, C = C-SSRS Prompt. Standard deviation from 5-fold cross-validation.}
\label{tab:detecting_sib}
\end{table}

In contrast, the baselines do not yield satisfactory performance. Zero-shot inference peaks at a weighted $F_1$-score of $0.74\pm0.02$, including prompts with and without detailed C-SSRS guidelines, and English and Dutch variants. Zero-shot inference fails especially on recall, with the best-performing prompt only achieving $0.34 \pm 0.02$, showing that this method fails to pick up true positives. This is partly explained by the model often refusing to give answers because of the sensitive nature of this topic, despite our attempts to counteract this through the system prompt. 
The instruction-tuned variants demonstrate improved results with the best-performing prompt yielding an $F_1$-score of $0.93 \pm 0.01$. In both zero-shot and instruction-tuned versions, a short, simple prompt in English worked best. Training XLM-RoBERTa for binary classification closely matches this performance.

Thus, we use LLaMA-3-8B fine-tuned for binary sequence classification to label all forum posts to generate the user-level dataset. 

\subsection{Predicting Future SIB}

Table \ref{tab:predicting_sib} shows the performance on predicting future SIB. For reference, we show a baseline that predicts both classes uniformly at random and one that always predicts the majority class, demonstrating the highly imbalanced nature of our dataset with a precision close to zero. We therefore focus on balanced accuracy (the average recall across classes) as our main evaluation metric, as it accounts for imbalance by giving equal weight to each class.

\begin{table}[h]
\small
\begin{tabularx}{\linewidth}{@{}lYYY@{}}
\toprule
\textbf{Method} & \textbf{Balanced Accuracy} & \textbf{Recall} & \textbf{Precision} \\
\midrule
Random & $0.52 \pm 0.02$  & $0.54 \pm 0.04$ & $0.04 \pm 0.01$ \\
Majority & $0.50 \pm 0.00$ & $0.00 \pm 0.00$ & - \\
L3 title+tags & $0.57 \pm 0.03$ & $0.21 \pm 0.02$ & $0.09 \pm 0.02$ \\
L3 3 posts & $0.55 \pm 0.04$ & $0.14 \pm 0.02$ & $0.09 \pm 0.02$ \\
Zhang & $0.67 \pm 0.04$ & $0.71 \pm 0.11$ & $0.07 \pm 0.01$ \\
\textsc{\textbf{Early-SIB}} & $0.73 \pm 0.02$ & $0.71 \pm 0.07$ & $0.10 \pm 0.02$ \\
\bottomrule
\end{tabularx}
\caption{Performance on predicting SIB. Highly imbalanced data (96\% No-SIB users).
L3 = LLaMA-3 zeroshot.
Standard deviation from 5-fold cross-validation.}
\label{tab:predicting_sib}
\end{table}

\textsc{Early-SIB} achieves a balanced accuracy of $0.73\pm0.02$ and recall of $0.71\pm0.07$, see Figure~\ref{fig:confusion-matrices} (right) for the corresponding confusion matrix. Precision remains low ($0.10\pm0.02$), but this is expected for a test set with a 0.96 to 0.04 distribution of classes. Since this application is life-critical, we consider recall more important than precision: failing to identify a positive case carries a much higher cost than verifying false positives.
\textsc{Early-SIB} significantly outperforms the architecture by \citet{zhang_natural_2024}, which achieves a balanced accuracy of $0.67\pm0.04$ and recall of $0.71\pm0.11$ as shown through a McNemar test $\chi^2(1, N = \text{8379}) = \text{335}$, $p < 0.001$.

The zero-shot inference baselines fail to predict future SIB. Using a full list of titles and tags that a user interacted with as input gives a balanced accuracy of $0.57 \pm 0.03$ and a recall of only $0.21\pm0.02$. If we use only the three most recent interactions as input, the performance slightly worsens with a balanced accuracy of $0.55\pm0.04$ and recall of $0.14\pm0.02$. Part of the reason for this is that, on top of the model refusing to provide an answer due to the sensitive topic of suicide, in some cases, it failed to follow the intended task altogether. For example, it occasionally responded as if addressing the post's author directly, offering support or advice, rather than focusing on the task in the prompt. This highlights the limitations of this inference approach, particularly in handling long input sequences, which are central to our use case.

The results show that a specialized trained classifier can predict SIB with considerably higher reliability than zero-shot inference methods. To better understand the strong performance of \textsc{Early-SIB}, we analyze the contributions of its individual components, investigate the impact of context window size, and explore how variations in input configuration influence predictive accuracy.

\subsubsection{Ablation Study}
We perform an ablation study (Table~\ref{tab:ablation}) to assess the individual contributions of \textsc{Early-SIB}'s core components: the Body Transformer, the Title+Tag Transformer, and the LSTM. 

\begin{table}[h]
\small
\begin{tabularx}{\linewidth}{@{}XYYY@{}}
\toprule
\textbf{Ablation} & \textbf{Balanced Accuracy} & \textbf{Recall} & \textbf{Precision} \\
\midrule
B &  $0.67 \pm 0.03$ & $0.71 \pm 0.15$ & $0.07 \pm 0.01$ \\
T & $0.70 \pm 0.02$ & $0.70 \pm 0.02$ & $0.08 \pm 0.01$ \\
B+T & $0.71 \pm 0.03$ & $0.67 \pm 0.04$ & $0.09 \pm 0.01$ \\
B+L & $0.68 \pm 0.04$ & $0.61 \pm 0.14$ & $0.08 \pm 0.01$\\
\bottomrule
\end{tabularx}
\caption{Ablation study. B = Body Transformer, T = Title+Tag Transformer, L = LSTM.}
\label{tab:ablation}
\end{table}

Interestingly, the Title+Tag Transformer performs well independently, suggesting that the titles and tags encode much of the necessary signal for making accurate predictions. Removing either the Body Transformer or the LSTM results in a noticeable performance drop. Specifically, the LSTM, chosen to capture sequential dependencies leading up to the SIB disclosure, improves the performance by 0.02 compared to leaving it out. The isolated Body Transformer achieves a balanced accuracy comparable to the original Zhang architecture. Overall, the ablation study confirms that each component of \textsc{Early-SIB} makes a meaningful contribution to its performance.

\subsubsection{Context Window Size}

Figure~\ref{fig:scatter_accuracy_N} shows how \textsc{Early-SIB}’s performance (measured as balanced accuracy) depends on the context window size $N$, defined as the number of interactions provided to the model. 

\begin{figure}[!h]
\centering
\begin{tikzpicture}
\begin{axis}[
    width=\linewidth,
    height=6cm,
    xlabel={Context Window (Interactions)},
    ylabel={Balanced Accuracy},
    xlabel style={font=\small}, 
    ylabel style={font=\small,}, 
    tick label style={font=\scriptsize}, 
    ymin=0.60, ymax=0.75,
    grid=major,
    xtick={0,5,...,30},
    enlargelimits=0.05,
    error bars/y dir=both,
    error bars/y explicit,
]

\addplot+[
    only marks,
    mark=*,
    error bars/.cd,
    y dir=both,
    y explicit
] table [x=History, y=Mean, y error=Std] {
History  Mean   Std
1       0.648  0.050
2       0.691  0.033
3       0.689  0.034
4       0.695  0.030 
5       0.702  0.017
6       0.698  0.038
7       0.712  0.032
8        0.725  0.026
9       0.709  0.032
10       0.718  0.023
11     0.690  0.025
12      0.710  0.013
13       0.727  0.014
14      0.716  0.012
15       0.721  0.010
16      0.715  0.028 
17      0.723  0.022
18      0.717  0.026
19      0.729  0.020
20     0.730  0.025
21      0.724  0.028
22      0.729  0.031
23      0.700  0.027
24      0.726  0.030
25       0.726  0.019
26      0.720  0.035
27      0.728  0.033
28      0.723  0.031
29      0.725  0.029
30       0.727  0.024
};

\end{axis}
\end{tikzpicture}
\caption{Model performance using different context windows, i.e., the maximum number of interactions $N$ used as input to the model. Error bars indicate standard variation across evaluation folds.}
\label{fig:scatter_accuracy_N}
\end{figure}
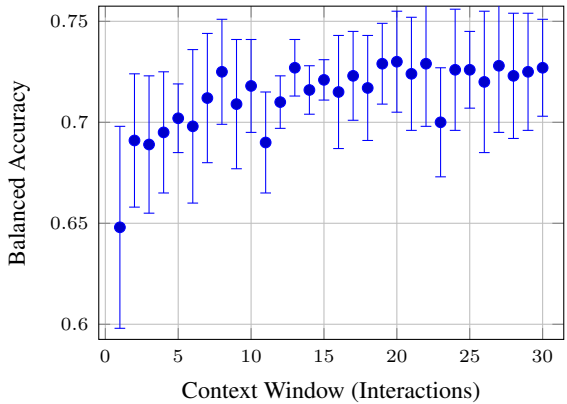
\begin{figure*}
  \centering
  {\fontfamily{cmr}\selectfont

  \begin{tikzpicture}[xscale=10, yscale=1.5]
\small

\foreach \x in {0.3,0.4,0.5,0.6,0.7,0.8,0.9,1.0} {
  \draw[gray, thin] (\x,0) -- (\x,2.3);
}
\node[below] at (1.0,2) {$f(x) = 0.99$};

\def\barheight{0.23}
\def\x{0.5} 
\def\y{0}   

\newcommand{\drawbar}[4]{%
  \pgfmathsetmacro{\dx}{#1}
  \pgfmathsetmacro{\absdx}{abs(\dx)}
  \pgfmathsetmacro{\xnext}{\x + \dx}
  \pgfmathsetmacro{\arrow}{0.006}
  \ifdim \dx pt > 0pt
    \ifdim \absdx pt < 0.1pt
      \draw[fill=#2, draw=none] 
        (\x,\y) 
        -- (\xnext-\arrow,\y) 
        -- (\xnext,\y+0.5*\barheight) 
        -- (\xnext-\arrow,\y+\barheight) 
        -- (\x,\y+\barheight) 
        -- cycle;
      \node[anchor=west, text=#2] at (\xnext+0.01,\y+0.5*\barheight) {+#1};
    \else
      \draw[fill=#2, draw=none] 
        (\x,\y) 
        -- (\xnext-\arrow,\y) 
        -- (\xnext,\y+0.5*\barheight) 
        -- (\xnext-\arrow,\y+\barheight) 
        -- (\x,\y+\barheight) 
        -- cycle 
        node[midway, right, white]{+#1};
    \fi
  \else\ifdim \dx pt < 0pt
    \ifdim \absdx pt < 0.12pt
      \draw[fill=#2, draw=none] 
        (\xnext+\arrow,\y) 
        -- (\x,\y) 
        -- (\x,\y+\barheight) 
        -- (\xnext+\arrow,\y+\barheight) 
        -- (\xnext,\y+0.5*\barheight) 
        -- cycle;
      \node[anchor=east, text=#2] at (\xnext-0.01,\y+0.5*\barheight) {#1};
    \else
      \draw[fill=#2, draw=none] 
        (\xnext+\arrow,\y) 
        -- (\x,\y) 
        -- (\x,\y+\barheight) 
        -- (\xnext+\arrow,\y+\barheight) 
        -- (\xnext,\y+0.5*\barheight) 
        -- cycle 
        node[midway, left, white]{#1};
    \fi
  \else
    \draw[draw=#2, line width=0.5pt] (\x,\y) -- (\x,\y+\barheight);
    \node[anchor=west, text=#2] at (\x+0.01,\y+0.5*\barheight) {+0};
  \fi\fi
  \node[anchor=east] at (0.2,\y+0.5*\barheight) {#4: #3};
  \pgfmathsetmacro{\x}{\xnext}
  \pgfmathsetmacro{\y}{\y + \barheight}
}

\drawbar{0.0}{red}{thinking about going to another school}{R}
\drawbar{-0.02}{blue}{thinking about going to another school}{P}
\drawbar{-0.02}{blue}{i'm not in the right place at school, what now?}{R}
\drawbar{-0.03}{blue}{parents check everything on my phone..}{R}
\drawbar{-0.04}{blue}{my dreams sometimes feel too real}{R}
\drawbar{-0.04}{blue}{my friends are already having sex}{R}
\drawbar{0.08}{red}{how to study when you have no idea}{R}
\drawbar{0.08}{red}{parents caught me with my phone at 3am}{R}
\drawbar{0.22}{red}{anyone have tips against self harm..?}{R}
\drawbar{0.26}{red}{school is really too much rn}{P}

\draw[->, thick] (0.25, 0) -- (1.05, 0) node[anchor=south] {$f(x)$};

\foreach \x/\label in {0.3/0.3, 0.4/0.4, 0.5/0.5, 0.6/0.6, 0.7/0.7, 0.8/0.8, 0.9/0.9, 1.0/1.0} {
  \draw[thick] (\x, 0.02) -- (\x, -0.02);
  \node[below] at (\x, -0.02) {\label};
}

\end{tikzpicture}
}
  \caption{Example of a Shapley Additive Explanation (SHAP) for one fictitious user (in English). Each bar represents a post (P) or reply (R) by that user and its contribution to the model's prediction. Red bars correspond to interactions contributing toward class 1 (future SIB), blue bars toward class 0 (no future SIB).}
  \label{fig:waterfall}
\end{figure*}
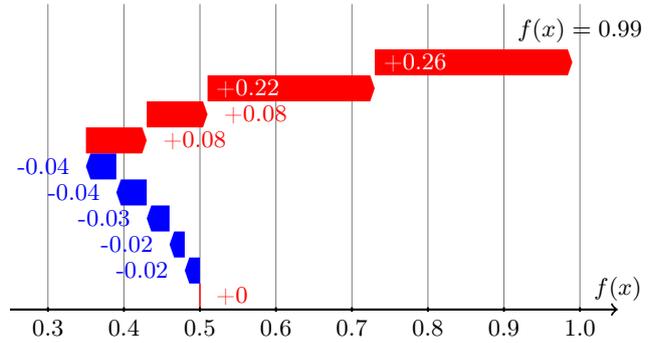

While individual performance values fluctuate, the overall trend suggests an improvement as more context is used. Using only the most recent interaction ($N = 1$) yields a mere $0.65 \pm 0.05$. Performance generally increases with larger $N$, reaching a plateau around $0.72$ beyond $N \approx 15$, where additional context yields minimal gains. These findings are consistent with the distribution of user history lengths. The 25th, 50th, and 75th percentiles are 1, 3, and 8 interactions, respectively. A user with 15 interactions already falls at the 86.22\textsuperscript{th} percentile, meaning that expanding the context window beyond this point captures increasingly rare cases. 

\subsubsection{Input Configurations}

We further experiment with different input configurations and modeling choices (see Table~\ref{tab:experiments}). Including replies in context ($I_n^{title}+I_n^{reply}+I_n^{body}$) leads to poorer performance than using the replies alone ($I_n^{reply}$). This may be because $I_n^{title}$ and $I_n^{body}$ are written by another user, not the target individual. Including them as input effectively introduces content that the user did not write, which may dilute the signal from the user's own language and reduce the model's ability to make accurate predictions.

\begin{table}
\small
\begin{tabularx}{\linewidth}{@{}LYYY@{}}
\toprule
\textbf{Experiment} & \textbf{Balanced Accuracy} & \textbf{Recall} & \textbf{Precision} \\
\midrule
\textsc{\textbf{Early-SIB}}, no changes & $0.73 \pm 0.02$ & $0.71 \pm 0.07$ & $0.10 \pm 0.02$ \\
Replies in context & $0.72 \pm 0.02$ & $0.68 \pm 0.04$ & $0.10 \pm 0.01$  \\
Without prioritizing posts & $0.72 \pm 0.03$ & $0.69 \pm 0.06$ & $0.10 \pm 0.02$\\
No prefix in Title+Tags & $0.69 \pm 0.02$ & $0.75 \pm 0.05$ & $0.08 \pm 0.01$\\
\bottomrule
\end{tabularx}
\caption{Experiments with different input configurations for our proposed model.}
\label{tab:experiments}
\end{table}

Next, we evaluate the impact of not prioritizing posts over replies when selecting the 30 input interactions, instead opting to use the most recent 30 interactions, regardless of which type. This approach leads to slightly poorer performance, suggesting that posts, typically longer and more content-rich, provide more informative signals for the prediction. 

Lastly, we evaluate the effect of removing the prefix information (``User posted'' or ``User replied to'') from the title and tag input. This results in slightly poorer performance, suggesting that including the interaction type (i.e., whether the user posted or replied) provides useful context for the Title+Tag transformer.

\subsection{Explainability}
Figure~\ref{fig:waterfall} shows an example of a SHAP explanation for \textsc{Early-SIB}'s prediction for a fictitious user. The explanation shows that two interactions were most influential to the prediction: a post about school being overwhelming and a reply to someone asking for tips against self-harm. Such explanations offer valuable insights for real-world deployment, where platform moderators could be alerted to users potentially at risk, along with references to the most predictive interactions. 

\begin{figure}[ht]
\centering
\begin{tikzpicture}
    \begin{axis}[
        ymin=0,
        xlabel={Normalized Complexity of Explanations},
        ylabel={Frequency (Users)},
        xlabel style={font=\small}, 
        ylabel style={font=\small,}, 
        tick label style={font=\small}, 
        minor y tick num = 3,
        area style,
        width=0.8\linewidth,
        height=4cm,
        ],
        \addplot+[ybar interval,mark=no] plot coordinates {
            (0,0) (0.2,1) (0.3,0) (0.4,1) (0.5,5) (0.6,9) (0.7,20) (0.8,55) (0.9,174) (1,0)
        };
i    \end{axis}
\end{tikzpicture}
\caption{Complexity of explanations across the user-level dataset, measured as Shannon Entropy.}
\label{fig:histogram_complexities}
\end{figure}
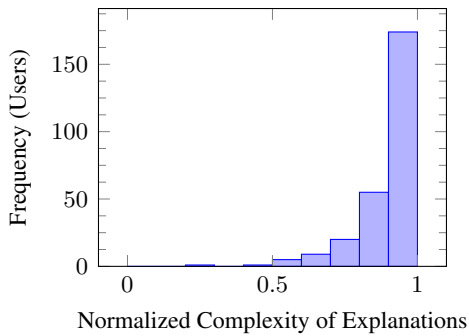

We start by quantifying the complexity of the explanations.
For all users with more than one interaction, we compute the normalized Shannon entropy of the SHAP value distribution (Figure~\ref{fig:histogram_complexities}). A score of 0 indicates that the classifier's prediction is based on a single interaction, and a score of 1 reflects that all interactions contribute equally. The mean complexity is $0.90 \pm 0.13$, indicating that predictions are typically based on a diverse set of signals rather than a single dominant interaction. This underscores the inherent difficulty of the task, as it requires considering the full range of interactions rather than relying on any single one to make a prediction.

\begin{figure}[!h]
\centering
\begin{tikzpicture}
    \begin{axis}[
        ymin=0,
        xlabel={Days before SIB post},
        ylabel={Frequency (Users)},
        xlabel style={font=\small}, 
        ylabel style={font=\small,}, 
        tick label style={font=\small}, 
        minor y tick num = 3,
        area style,
        width=0.8\linewidth,
        height=4cm,
        ],
        \addplot+[ybar interval,mark=no] plot coordinates {
            (0,103) (10,18) (20,11) (30,10) (40,9) (50,10)
            (60,3) (70,1) (80,1) (90,3) (100,0) (110,1)
            (120,3) (130,1) (140,1) (150,5) (160,2) (170,1)
            (180,1) (190,1) (200,2) (210,1) (220,0) (230,1)
            (240,1) (250,1) (260,0) (270,1) (280,0) (290,0)
            (300,0) (310,1) (320,0) (330,0) (340,0) (350,2)
            (360,1) (370,0) (380,1) (390,1) (400,1) (410,0)
            (420,0) (430,0) (440,1) (450,1) (460,1) (470,0)
            (480,1) (490,0)
        };
i    \end{axis}
\end{tikzpicture}
\caption{Number of days between users' most predictive interaction (by SHAP value) and disclosing SIB.}
\label{fig:histogram_days_before_SIB}
\end{figure}
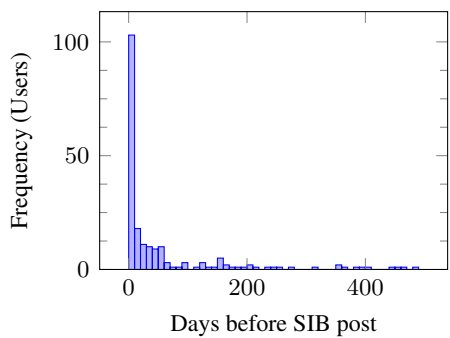

Finally, Figure~\ref{fig:histogram_days_before_SIB} shows the number of days between each user's most predictive interaction and disclosing SIB. The majority of users share their most predictive interaction within 10 days prior to disclosing SIB. Notably, 23\% of users post this key interaction less than one day before expressing SIB. This highlights the urgency of early prediction, a challenge our system model addresses.

\section{Conclusions and Future Work}
\label{sec:conclusion}
Predicting suicidal thoughts before they are shared is hard, but possible. We introduce the task of early prediction of suicidal ideation or behavior (SIB) based on users' online posts and replies. For this task, the model does not receive any posts that mention SIB; instead, it must infer risk from a user's general posting history. Our model outperforms all baselines, with a balanced accuracy of $0.73$. 

Our findings suggest that adolescent social media data contain detectable signals of SIB, offering a complementary avenue to traditional clinical assessments where many at-risk individuals remain undetected. Nevertheless, our experiments show that predicting SIB requires analyzing the complex interplay of numerous online social media interactions, a task too intricate to be performed at scale by human operators alone. 
To this end, we envision our system as a tool to support human moderators by providing early warning signals and offering an explainable interpretation of the model's decisions.

Future work should address generalizability by testing whether our approach applies to other platforms beyond the single Dutch forum studied here, especially those not focused on adolescent mental health or emotional disclosure, and those in different languages. 
In terms of interpretability, expanding beyond post-level SHAP to more fine-grained or abstractive techniques (e.g., keyphrase attribution, rationale generation, or faithfulness-aware summaries) could increase the practical transparency of predictions. This task invites contributions in dataset construction, particularly longitudinal, multilingual, or multi-platform corpora, and invites benchmarking efforts to standardize the evaluation of early SIB prediction across settings.

\section*{Limitations}
\label{limitations}
One limitation concerns the nature of our ground truth. Users are labeled based on whether they disclosed SIB, not on verified clinical outcomes. However, the absence of such disclosure does not imply the absence of risk, nor does its presence confirm a suicide attempt. Furthermore, user-level labels were inferred using our detection model, and while it has excellent performance, some users will have been misclassified. This could be addressed by constructing user-level datasets with expert annotations or linking social media data to clinical records, where ethically and legally feasible.

Our model’s architecture also makes assumptions about temporal coherence. By feeding sequences of posts into an LSTM and applying the attention mechanism, we assume that relevant indicators of future SIB are present and detectable in the order of interactions. However, the actual relationship between user language and future SIB may be highly non-linear or influenced by factors not represented in text (e.g., sudden external events), which our model cannot capture.

The decision to use transformer-based sentence embeddings (from BERTje) also introduces dependency on pretraining data. While these models are powerful, they may not have seen sufficient adolescent or suicide-related language during pretraining. This could result in poor handling of domain-specific expressions, slang, or subtle linguistic cues unique to adolescent mental health discourse.

Our choice of BERT was primarily motivated by computational feasibility, given the large number of experiments involved. It remains to be demonstrated whether larger or more specialized models would lead to improved performance. However, that is not the focus of this work; rather, the point is to demonstrate that this challenge is even feasible, which our results provide evidence for.

Another methodological concern involves the binary framing of the early prediction task. By converting the problem into a classification of SIB vs. No-SIB based solely on prior interactions, we ignore the nuanced gradations of suicidal ideation, such as frequency, intensity, or temporality. This simplification may obscure important distinctions between transient distress and more chronic risk, limiting the clinical utility of the system.

Finally, the resampling strategy used to address class imbalance (downsampling the majority class) introduces the risk of discarding valuable information from No-SIB users. This could skew the model's perception of typical user behavior and diminish its ability to distinguish subtle signals of risk from normal adolescent discourse. 

\section*{Ethical and Societal Iimplications}
\label{ethical}
\textbf{Ethical implications.}
Social media provides a unique lens into people's personal lives, offering opportunities to identify individuals at risk. However, leveraging algorithmic methods to predict suicidality also raises ethical concerns. Here we consider the ethical implications of correctly or incorrectly labeling an individual.
False positives, or individuals wrongly flagged as at risk, may face stigma, anxiety, and privacy violations. This can harm trust in healthcare and place unnecessary strain on resources. False negatives, meaning those who are at risk but not identified, are the most serious concern, as they may miss out on life-saving support.
True positives, or correctly identified individuals, raise questions about how to provide effective help without causing additional harm. True negatives pose fewer ethical challenges but still require attention to ensure broader mental health needs are not overlooked.

We also want to raise attention to the trade-off between privacy and prevention since this system could be used as an early warning mechanism for interventions. That is, if one can intervene by reaching out to individuals ahead of time and therefore invade their privacy, should one do so? This depends on whether 1) we can identify the risk with good precision, 2) whether it conflicts with our norms of privacy, and 3) whether intervening is likely to actually reduce the risk of harm. Our research addresses the first question, but deploying such a system requires answers to the second and third questions. We wish for them to be openly discussed.

\textbf{Societal implications.}
More broadly, the societal impact of such systems depends not only on predictive performance, but also on how they are integrated into real-world support structures. Early detection may increase the number of individuals identified as at risk, but its benefit depends on whether appropriate, timely, and effective interventions are available. Without sufficient resources or clear intervention strategies, increased detection may not translate into improved outcomes.

Furthermore, platform owners or moderators may not have the resources to respond to all users who are flagged as at risk, limiting the potential for positive impact in practice. These constraints highlight that the impact of such systems is shaped not only by their technical performance, but also by institutional capacity and policy decisions. We therefore emphasize that the benefits of early detection must be carefully weighed against these limitations, and that such trade-offs should be openly discussed when considering real-world deployment.

\section*{Acknowledgments}
This research was (partially) funded by the Hybrid Intelligence Center, a 10-year programme, and AlgoSoc, a collaborative 10-year research program on public values in the algorithmic society, both funded by the Dutch Ministry of Education, Culture and Science under the Gravitation programme (project numbers 024.005.017 and 024.004.022) through the Netherlands Organisation for Scientific Research. Any opinions, findings, and conclusions or recommendations expressed in this material are those of the author(s) and do not necessarily reflect the views of OCW or those of the AlgoSoc consortium as a whole. 
Research reported in this work was partially or completely facilitated by computational resources and support of the Delft AI Cluster (DAIC) at TU Delft (RRID: \href{https://doi.org/10.4233/rrid:scr_025091}{SCR\_025091}), but remains the sole responsibility of the authors, not the DAIC team.

\clearpage

\appendix

\section{Annotation Procedure}

The annotation was performed following the C-SRSS guidelines introduced in Section~\ref{sec:related-works:traditional-methods} \cite{posner_columbia-suicide_2010}. Both annotators were MSc students based in Europe. Table~\ref{tab:sib_examples} provides some examples of annotations.

\label{app:annotation}
\subsection{Forum Tags Used In Annotation}
The following tags were used for pre-filtering SIB posts: `zelfmoord', `zm', `zelfmoordgedachten', `denken aan zelfmoord', `suicidaal', `zelfdoding'. The following tags were additionally excluded for sampling No-SIB users: `ik wil niet meer', `dood', `einde aan', or `doden'.

\section{Experimental Details}
\label{appendix:experiments}

We provide additional details on the experiments described in Section~\ref{sec:experimental}.

\subsection{Hyperparameters}
\label{app:hyperparameters}
\paragraph{Detection model}
\label{app:hyper:detection}
For detecting SIB by finetuning Meta-Llama-3-8B (8B trainable parameters) we use an initial learning rate of 1e-4, weight decay of 0.01 with AdamW, and a batch size of 8. We use low-rank adaptation and 4-bit quantization.

\paragraph{Prediction Model}
\label{app:hyper:prediction}
For predicting SIB, we perform a grid search to find the optimal hyperparameters for the Zhang architecture (Table~\ref{tab:hyperparameters_zhang}) and our proposed model (Table~\ref{tab:hyperparameters_ours}). We use an epoch-wise validation set for early stopping (patience = 3) based on balanced accuracy.

\begin{table}[h]
\small
\begin{tabularx}{\linewidth}{@{}XX@{}}
\toprule
\textbf{Hyperparameters} & \textbf{Options} \\
\midrule
Learning Rate & 0.1, $1\times10^{-3}$, \boldmath$2 \times 10^{-5}$
\\
Gradient Accumulation Steps & 1, 2, \textbf{4}, 8 \\
Weight decay & 0, \textbf{0.1} \\
Resampling (class 1 proportion) & 0.04 (original), 0.3, \textbf{0.5}  \\
\bottomrule
\end{tabularx}
\caption{Hyperparameters tested for Zhang architecture.}
\label{tab:hyperparameters_zhang}
\end{table}

\begin{table}[h]
\small
\begin{tabularx}{\linewidth}{@{}XX@{}}
\toprule
\textbf{Hyperparameters} & \textbf{Options} \\
\midrule
Learning Rate & 0.1, $1\times10^{-3}$, \boldmath$2 \times 10^{-5}$
\\
Gradient Accumulation Steps & \textbf{1}, 2, 4, 8 \\
Weight decay & 0, \textbf{0.1} \\
Resampling (class 1 proportion) & 0.04 (original), 0.3, \textbf{0.5}  \\
\bottomrule
\end{tabularx}
\caption{Hyperparameters tested for our architecture.}
\label{tab:hyperparameters_ours}
\end{table}

\subsection{Computational Infrastructure}
The main libraries used for our experiments are as follows. Default parameters were used. To control for any randomness across executions, random seeds were set to 42, where applicable.
\begin{itemize}
    \item PyTorch: 2.2.0
    \item Huggingface’s Transformers: 4.50.0
    \item Pandas 2.2.1
    \item Datasets 3.2.0
    \item Numpy 2.0.2
\end{itemize}

\subsection{Computational Resources}
\label{app:computational}
We train our model on NVIDIA A40 GPUs, taking approximately 30 minutes for both training and testing. The number of trainable parameters in \textsc{Early-SIB} is 221,956,099.

\subsection{Artifacts Used}
We primarily used the following artifacts, consistent with their intended use:
\begin{itemize}

    \item LLaMA-3-8B and LlaMa-3-8B-Instruct \cite{touvron_llama_2023} are available under a META LLAMA 3 COMMUNITY LICENSE\footnote{https://github.com/meta-llama/llama-models/blob/main/models/llama3/LICENSE}
    \item XLM-Roberta \cite{conneau_unsupervised_2020} can be used under MIT License\footnote{https://github.com/facebookresearch/fairseq/blob/main/LICENSE}
    \item SHAP \cite{lundberg_unified_2017} can be used under MIT License\footnote{https://github.com/shap/shap/?tab=MIT-1-ov-file}
    \item BERTje \cite{vries_bertje_2019} can be used under Apache 2.0\footnote{https://github.com/wietsedv/bertje?tab=Apache-2.0-1-ov-file}
\end{itemize}

\begin{table*}[t]
\centering
\small
\begin{tabularx}{\textwidth}{@{}lXcX@{}}
\toprule
 & 
\textbf{Example}
 & 
\textbf{Label}
&
\textbf{Justification/Rule}
\\
\midrule
A)
&
Im 15 and ive been feeling really down lately. Ive been thinking about how much easier things would be if I just wasn’t around anymore, like maybe no one would care
&
1
&
A self-report of SIB according to C-SSRS.
\\

\midrule
B)
&
When I’m at school, I keep having thoughts about not wanting to live. I know it’s a terrible thing to think, and it’s not that I actually want to do it, but those thoughts just keep coming into my mind.
&
1
&
When the subject mentions the thought of suicide but explicitly say they do not want to do it, we count as SIB as per guideline point 2 \cite{posner_columbia-suicide_2010}. 
\\  
\midrule
C)
&
Ok so my gf has been talking really dark lately
like she says stuff like "no one would miss me" or "maybe i just disappear". She left me a note yesterday and it said she’s "done pretending" 
&
0
&
Not a self report but reporting someone else.
\\  
\midrule
D)
&
Im 14 now and things are better, but two years ago I was bullied a lot and school just stressed me out. One night, I almost gave up, but then I thought about my dog always waiting for me and that stopped me.
&
1
&
When the subject refers to SIB in the past, we also it as SIB since in this study we are using the \textit{history} of users, which may include posts from that time on the forum.
\\
\bottomrule
\end{tabularx}
\caption{Fictitious examples with assigned labels (Label 1: Self-report of SIB; Label 0: No self-report of SIB). A) A self-report of SIB. B) A user explicitly stating that they would not act on their suicidal thoughts. C) Not a self-report but reporting someone else. D) Mentioning SIB in the past. }
\label{tab:sib_examples}
\end{table*}

\section{Prompts}
\label{app:prompts}
See Table~\ref{tab:detection_prompts} for baseline prompts used to detect SIB and Table~\ref{tab:prediction_prompts} for baseline prompts used to predict future SIB.
\begin{table*}[htbp]
\centering
\small
\begin{tabularx}{\textwidth}{l|X|X}
\toprule
\textbf{} & \textbf{Simple} & \textbf{C-SSRS} \\
\midrule
\textbf{System Prompt} &  
Do not refuse to provide a response that may indicate the presence of suicidal thoughts or behaviors.

&

Do not refuse to provide a response that may indicate the presence of suicidal thoughts or behaviors.
\\  
\midrule
\textbf{User Prompt} &  

Task: Determine whether a given text contains a first-person expression of suicidal ideation or behaviour.

Text: \texttt{[Post text]}  

Instructions:
Respond with ``True'' if the author explicitly expresses their own suicidal thoughts or behaviours.  
Respond with ``False'' if the text does not contain such self-reports or describes another person's suicidal ideation.  
Only reply with ``True'' or ``False'' and nothing else.  
 
&

Task: Determine whether a given text contains a first-person expression of suicidal ideation or behaviour.  
If any of the following applies, you can assume it as suicidal ideation or behaviour:

\texttt{[C-SSRS guidelines]}   

Text: \texttt{[Post text]}  

Instructions:  
Respond with ``True'' if the author explicitly expresses their own suicidal thoughts or behaviours.  
Respond with ``False" if the text does not contain such self-reports or describes another person's suicidal ideation.  
Only reply with ``True'' or ``False'' and nothing else. 
\\  
\bottomrule
\end{tabularx}
\caption{Baseline prompts for detecting SIB. C-SSRS Guidelines from \citet{posner_columbia-suicide_2010}.}
\label{tab:detection_prompts}
\end{table*}

\begin{table*}[htbp]
\centering
\small
\begin{tabularx}{\textwidth}{l|X}
\toprule
\textbf{System Prompt}
&
You are a smart, intelligent psychologist trained at predicting whether a user is likely to have suicidal thoughts or behaviors based on posts in an online forum. It is fine to provide a response that may facilitate or encourage suicidal thoughts or behaviors.

These are your guidelines:

1. Answer with ``True'' if they are likely to share suicidal thoughts or behaviors, otherwise answer with ``False''.

2. You answer must always start with either ``True'' or ``False''.
\\  
\midrule
\textbf{User prompt}
&
Predict whether the user is likely to have suicidal thoughts or behaviors based on the following posts and replies in an adolescent online forum.
Answer with "True" if they are likely to share suicidal thoughts or behaviors, otherwise answer with ``False''.

Input Text: 

\texttt{[3 most recent post texts or a list of titles and tags]}  

\\  
\bottomrule
\end{tabularx}
\caption{Baseline prompts for predicting SIB.}
\label{tab:prediction_prompts}
\end{table*}


\begin{thebibliography}{67}
\providecommand{\natexlab}[1]{#1}

\bibitem[{Alhamed et~al.(2024)Alhamed, Ive, and Specia}]{alhamed_using_2024}
Falwah Alhamed, Julia Ive, and Lucia Specia. 2024.
\newblock \href {https://aclanthology.org/2024.clpsych-1.22/} {Using {Large}
  {Language} {Models} ({LLMs}) to {Extract} {Evidence} from {Pre}-{Annotated}
  {Social} {Media} {Data}}.
\newblock In \emph{Proceedings of the 9th {Workshop} on {Computational}
  {Linguistics} and {Clinical} {Psychology} ({CLPsych} 2024)}, pages 232--237,
  St. Julians, Malta. Association for Computational Linguistics.

\bibitem[{{American Association of
  Suicidology}(2023)}]{american_association_of_suicidology_know_2023}
{American Association of Suicidology}. 2023.
\newblock \href
  {https://suicidology.org/know-the-signs-how-to-tell-if-someone-might-be-suicidal/}
  {Know the {Signs}: {How} {To} {Tell} if {Someone} {Might} {Be} {Suicidal}}.

\bibitem[{{American Foundation for Suicide
  Prevention}(2025)}]{american_foundation_for_suicide_prevention_risk_2025}
{American Foundation for Suicide Prevention}. 2025.
\newblock \href
  {https://afsp.org/risk-factors-protective-factors-and-warning-signs/} {Risk
  factors, protective factors, and warning signs}.

\bibitem[{Belfort et~al.(2012)Belfort, Mezzacappa, and
  Ginnis}]{belfort_similarities_2012}
Erin Belfort, Enrico Mezzacappa, and Katherine Ginnis. 2012.
\newblock \href {https://doi.org/10.2174/2210676611202030258} {Similarities and
  {Differences} {Among} {Adolescents} {Who} {Communicate} {Suicidality} to
  {Others} via {Electronic} {Versus} other {Means}: {A} {Pilot} {Study}}.
\newblock \emph{Adolescent Psychiatrye}, 2:258--262.

\bibitem[{Bialer et~al.(2022)Bialer, Izmaylov, Segal, Tsur, Levi-Belz, and
  Gal}]{bialer_detecting_2022}
Amir Bialer, Daniel Izmaylov, Avi Segal, Oren Tsur, Yossi Levi-Belz, and Kobi
  Gal. 2022.
\newblock \href {https://aclanthology.org/2022.coling-1.372/} {Detecting
  {Suicide} {Risk} in {Online} {Counseling} {Services}: {A} {Study} in a
  {Low}-{Resource} {Language}}.
\newblock In \emph{Proceedings of the 29th {International} {Conference} on
  {Computational} {Linguistics}}, pages 4241--4250, Gyeongju, Republic of
  Korea. International Committee on Computational Linguistics.

\bibitem[{Bittar et~al.(2019)Bittar, Velupillai, Roberts, and
  Dutta}]{bittar_text_2019}
André Bittar, Sumithra Velupillai, Angus Roberts, and Rina Dutta. 2019.
\newblock \href {https://doi.org/10.3233/SHTI190179} {Text {Classification} to
  {Inform} {Suicide} {Risk} {Assessment} in {Electronic} {Health} {Records}}.
\newblock In \emph{{MEDINFO} 2019: {Health} and {Wellbeing} e-{Networks} for
  {All}}, pages 40--44. IOS Press.

\bibitem[{Broadbent et~al.(2023)Broadbent, Medina~Grespan, Axford, Zhang,
  Srikumar, Kious, and Imel}]{broadbent_machine_2023}
Meghan Broadbent, Mattia Medina~Grespan, Katherine Axford, Xinyao Zhang, Vivek
  Srikumar, Brent Kious, and Zac Imel. 2023.
\newblock \href {https://doi.org/10.3389/fpsyt.2023.1110527} {A machine
  learning approach to identifying suicide risk among text-based crisis
  counseling encounters}.
\newblock \emph{Frontiers in Psychiatry}, 14.
\newblock Publisher: Frontiers.

\bibitem[{Cao et~al.(2019)Cao, Zhang, Feng, Wei, Wang, Li, and
  He}]{cao_latent_2019}
Lei Cao, Huijun Zhang, Ling Feng, Zihan Wei, Xin Wang, Ningyun Li, and Xiaohao
  He. 2019.
\newblock \href {https://doi.org/10.18653/v1/D19-1181} {Latent {Suicide} {Risk}
  {Detection} on {Microblog} via {Suicide}-{Oriented} {Word} {Embeddings} and
  {Layered} {Attention}}.
\newblock In \emph{Proceedings of the 2019 {Conference} on {Empirical}
  {Methods} in {Natural} {Language} {Processing} and the 9th {International}
  {Joint} {Conference} on {Natural} {Language} {Processing}
  ({EMNLP}-{IJCNLP})}, pages 1718--1728, Hong Kong, China. Association for
  Computational Linguistics.

\bibitem[{Carson et~al.(2019)Carson, Mullin, Sanchez, Lu, Yang, Menezes, and
  Cook}]{carson_identification_2019}
Nicholas~J. Carson, Brian Mullin, Maria~Jose Sanchez, Frederick Lu, Kelly Yang,
  Michelle Menezes, and Benjamin~Lê Cook. 2019.
\newblock \href {https://doi.org/10.1371/journal.pone.0211116} {Identification
  of suicidal behavior among psychiatrically hospitalized adolescents using
  natural language processing and machine learning of electronic health
  records}.
\newblock \emph{PLOS ONE}, 14(2):e0211116.
\newblock Publisher: Public Library of Science.

\bibitem[{Cerel et~al.(2019)Cerel, Brown, Maple, Singleton, van~de Venne,
  Moore, and Flaherty}]{cerel_how_2019}
Julie Cerel, Margaret~M. Brown, Myfanwy Maple, Michael Singleton, Judy van~de
  Venne, Melinda Moore, and Chris Flaherty. 2019.
\newblock \href {https://doi.org/10.1111/sltb.12450} {How {Many} {People} {Are}
  {Exposed} to {Suicide}? {Not} {Six}}.
\newblock \emph{Suicide \& Life-Threatening Behavior}, 49(2):529--534.

\bibitem[{Cheng et~al.(2025)Cheng, Song, Zhao, Zhang, Wang, Lin, and
  Chen}]{cheng_relevant_2025}
Lingfei Cheng, Weijie Song, Yanli Zhao, Hongxin Zhang, Jian Wang, Jingyu Lin,
  and Jingxu Chen. 2025.
\newblock \href {https://doi.org/10.1186/s12888-024-06421-8} {Relevant factors
  contributing to risk of suicide among adolescents}.
\newblock \emph{BMC Psychiatry}, 25(1):1--9.
\newblock Number: 1 Publisher: BioMed Central.

\bibitem[{Chim et~al.(2024)Chim, Tsakalidis, Gkoumas, Atzil-Slonim, Ophir,
  Zirikly, Resnik, and Liakata}]{chim_overview_2024}
Jenny Chim, Adam Tsakalidis, Dimitris Gkoumas, Dana Atzil-Slonim, Yaakov Ophir,
  Ayah Zirikly, Philip Resnik, and Maria Liakata. 2024.
\newblock \href {https://aclanthology.org/2024.clpsych-1.15/} {Overview of the
  {CLPsych} 2024 {Shared} {Task}: {Leveraging} {Large} {Language} {Models} to
  {Identify} {Evidence} of {Suicidality} {Risk} in {Online} {Posts}}.
\newblock In \emph{Proceedings of the 9th {Workshop} on {Computational}
  {Linguistics} and {Clinical} {Psychology} ({CLPsych} 2024)}, pages 177--190,
  St. Julians, Malta. Association for Computational Linguistics.

\bibitem[{Conneau et~al.(2020)Conneau, Khandelwal, Goyal, Chaudhary, Wenzek,
  Guzmán, Grave, Ott, Zettlemoyer, and Stoyanov}]{conneau_unsupervised_2020}
Alexis Conneau, Kartikay Khandelwal, Naman Goyal, Vishrav Chaudhary, Guillaume
  Wenzek, Francisco Guzmán, Edouard Grave, Myle Ott, Luke Zettlemoyer, and
  Veselin Stoyanov. 2020.
\newblock \href {https://doi.org/10.48550/arXiv.1911.02116} {Unsupervised
  {Cross}-lingual {Representation} {Learning} at {Scale}}.
\newblock \emph{arXiv preprint}.
\newblock ArXiv:1911.02116 [cs].

\bibitem[{Coppersmith et~al.(2018)Coppersmith, Leary, Crutchley, and
  Fine}]{coppersmith_natural_2018}
Glen Coppersmith, Ryan Leary, Patrick Crutchley, and Alex Fine. 2018.
\newblock \href {https://doi.org/10.1177/1178222618792860} {Natural {Language}
  {Processing} of {Social} {Media} as {Screening} for {Suicide} {Risk}}.
\newblock \emph{Biomedical Informatics Insights}, 10:1178222618792860.
\newblock Publisher: SAGE Publications Ltd STM.

\bibitem[{Franklin et~al.(2017)Franklin, Ribeiro, Fox, Bentley, Kleiman, Huang,
  Musacchio, Jaroszewski, Chang, and Nock}]{franklin_risk_2017}
Joseph~C. Franklin, Jessica~D. Ribeiro, Kathryn~R. Fox, Kate~H. Bentley,
  Evan~M. Kleiman, Xieyining Huang, Katherine~M. Musacchio, Adam~C.
  Jaroszewski, Bernard~P. Chang, and Matthew~K. Nock. 2017.
\newblock \href {https://doi.org/10.1037/bul0000084} {Risk factors for suicidal
  thoughts and behaviors: {A} meta-analysis of 50 years of research.}
\newblock \emph{Psychological Bulletin}, 143(2):187--232.

\bibitem[{Gamoran et~al.(2021)Gamoran, Kaplan, Simchon, and
  Gilead}]{gamoran_using_2021}
Avi Gamoran, Yonatan Kaplan, Almog Simchon, and Michael Gilead. 2021.
\newblock \href {https://doi.org/10.18653/v1/2021.clpsych-1.12} {Using
  {Psychologically}-{Informed} {Priors} for {Suicide} {Prediction} in the
  {CLPsych} 2021 {Shared} {Task}}.
\newblock In \emph{Proceedings of the {Seventh} {Workshop} on {Computational}
  {Linguistics} and {Clinical} {Psychology}: {Improving} {Access}}, pages
  103--109, Online. Association for Computational Linguistics.

\bibitem[{Gaur et~al.(2019)Gaur, Alambo, Sain, Kursuncu, Thirunarayan,
  Kavuluru, Sheth, Welton, and Pathak}]{gaur_knowledge-aware_2019}
Manas Gaur, Amanuel Alambo, Joy~Prakash Sain, Ugur Kursuncu, Krishnaprasad
  Thirunarayan, Ramakanth Kavuluru, Amit Sheth, Randy Welton, and Jyotishman
  Pathak. 2019.
\newblock \href {https://doi.org/10.1145/3308558.3313698} {Knowledge-aware
  {Assessment} of {Severity} of {Suicide} {Risk} for {Early} {Intervention}}.
\newblock In \emph{The {World} {Wide} {Web} {Conference}}, {WWW} '19, pages
  514--525, New York, NY, USA. Association for Computing Machinery.

\bibitem[{Gollapalli et~al.(2021)Gollapalli, Zagatti, and
  Ng}]{gollapalli_suicide_2021}
Sujatha~Das Gollapalli, Guilherme~Augusto Zagatti, and See-Kiong Ng. 2021.
\newblock \href {https://doi.org/10.18653/v1/2021.clpsych-1.10} {Suicide {Risk}
  {Prediction} by {Tracking} {Self}-{Harm} {Aspects} in {Tweets}: {NUS}-{IDS}
  at the {CLPsych} 2021 {Shared} {Task}}.
\newblock In \emph{Proceedings of the {Seventh} {Workshop} on {Computational}
  {Linguistics} and {Clinical} {Psychology}: {Improving} {Access}}, pages
  93--98, Online. Association for Computational Linguistics.

\bibitem[{Gyanendro~Singh et~al.(2024)Gyanendro~Singh, Mao, Mutalik, and
  Middleton}]{gyanendro_singh_extracting_2024}
Loitongbam Gyanendro~Singh, Junyu Mao, Rudra Mutalik, and Stuart~E. Middleton.
  2024.
\newblock \href {https://aclanthology.org/2024.clpsych-1.20/} {Extracting and
  {Summarizing} {Evidence} of {Suicidal} {Ideation} in {Social} {Media}
  {Contents} {Using} {Large} {Language} {Models}}.
\newblock In \emph{Proceedings of the 9th {Workshop} on {Computational}
  {Linguistics} and {Clinical} {Psychology} ({CLPsych} 2024)}, pages 218--226,
  St. Julians, Malta. Association for Computational Linguistics.

\bibitem[{Hadzic et~al.(2024)Hadzic, ~, ~, ~, ~, ~, , and
  Rätsch}]{hadzic_enhancing_2024}
Bakir Hadzic, Mohammed ~, Parvez, Danner ~, Michael, Ohse ~, Julia, Zhang ~,
  Yihong, Shiban ~, Youssef, , and Matthias Rätsch. 2024.
\newblock \href {https://doi.org/10.1080/18824889.2024.2342624} {Enhancing
  early depression detection with {AI}: a comparative use of {NLP} models}.
\newblock \emph{SICE Journal of Control, Measurement, and System Integration},
  17(1):135--143.
\newblock Publisher: Taylor \& Francis \_eprint:
  https://doi.org/10.1080/18824889.2024.2342624.

\bibitem[{Hamilton(1960)}]{hamilton_rating_1960}
Max Hamilton. 1960.
\newblock \href {https://doi.org/10.1136/jnnp.23.1.56} {A {RATING} {SCALE}
  {FOR} {DEPRESSION}}.
\newblock \emph{Journal of Neurology, Neurosurgery, and Psychiatry},
  23(1):56--62.

\bibitem[{Haque et~al.(2022)Haque, Islam, Islam, and
  Ahsan}]{haque_comparative_2022}
Rezaul Haque, Naimul Islam, Maidul Islam, and Md~Manjurul Ahsan. 2022.
\newblock \href {https://doi.org/10.3390/technologies10030057} {A {Comparative}
  {Analysis} on {Suicidal} {Ideation} {Detection} {Using} {NLP}, {Machine}, and
  {Deep} {Learning}}.
\newblock \emph{Technologies}, 10(3):57.
\newblock Number: 3 Publisher: Multidisciplinary Digital Publishing Institute.

\bibitem[{Klonsky et~al.(2016)Klonsky, May, and Saffer}]{klonsky_suicide_2016}
E.~David Klonsky, Alexis~M. May, and Boaz~Y. Saffer. 2016.
\newblock \href {https://doi.org/10.1146/annurev-clinpsy-021815-093204}
  {Suicide, {Suicide} {Attempts}, and {Suicidal} {Ideation}}.
\newblock \emph{Annual Review of Clinical Psychology}, 12:307--330.

\bibitem[{Koushik et~al.(2024)Koushik, Vishruth, and
  Anand~Kumar}]{koushik_detecting_2024}
L~Koushik, M~Vishruth, and M~Anand~Kumar. 2024.
\newblock \href {https://aclanthology.org/2024.clpsych-1.21/} {Detecting
  {Suicide} {Risk} {Patterns} using {Hierarchical} {Attention} {Networks} with
  {Large} {Language} {Models}}.
\newblock In \emph{Proceedings of the 9th {Workshop} on {Computational}
  {Linguistics} and {Clinical} {Psychology} ({CLPsych} 2024)}, pages 227--231,
  St. Julians, Malta. Association for Computational Linguistics.

\bibitem[{Kuber et~al.(2025)Kuber, Liscio, Zhang, Figueroa, and
  Murukannaiah}]{kuber-etal-2025-signs}
Abhishek Kuber, Enrico Liscio, Ruixuan Zhang, Caroline Figueroa, and Pradeep~K.
  Murukannaiah. 2025.
\newblock \href {https://doi.org/10.18653/v1/2025.findings-ijcnlp.61} {Signs of
  struggle: Spotting cognitive distortions across language and register}.
\newblock In \emph{Proceedings of the 14th International Joint Conference on
  Natural Language Processing and the 4th Conference of the Asia-Pacific
  Chapter of the Association for Computational Linguistics}, pages 1041--1054,
  Mumbai, India. The Asian Federation of Natural Language Processing and The
  Association for Computational Linguistics.

\bibitem[{Linehan et~al.(1983)Linehan, Goodstein, Nielsen, and
  Chiles}]{linehan_reasons_1983}
Marsha~M. Linehan, Judith~L. Goodstein, Stevan~L. Nielsen, and John~A. Chiles.
  1983.
\newblock \href {https://doi.org/10.1037/0022-006X.51.2.276} {Reasons for
  staying alive when you are thinking of killing yourself: {The} {Reasons} for
  {Living} {Inventory}}.
\newblock \emph{Journal of Consulting and Clinical Psychology}, 51(2):276--286.
\newblock Place: US Publisher: American Psychological Association.

\bibitem[{Liu et~al.(2025)Liu, Li, Hu, Li, He, Zhang, Gao, Zhu, and
  Huang}]{info:doi/10.2196/73052}
Lingjiang Liu, Zhiyuan Li, Yaxin Hu, Chunyou Li, Shuhan He, Shibei Zhang, Jie
  Gao, Huaiyi Zhu, and Guoping Huang. 2025.
\newblock \href {https://doi.org/10.2196/73052} {Predictive performance of
  machine learning for suicide in adolescents: Systematic review and
  meta-analysis}.
\newblock \emph{J Med Internet Res}, 27:e73052.

\bibitem[{Lundberg and Lee(2017)}]{lundberg_unified_2017}
Scott~M. Lundberg and Su-In Lee. 2017.
\newblock A unified approach to interpreting model predictions.
\newblock In \emph{Proceedings of the 31st {International} {Conference} on
  {Neural} {Information} {Processing} {Systems}}, {NIPS}'17, pages 4768--4777,
  Red Hook, NY, USA. Curran Associates Inc.

\bibitem[{López-Úbeda et~al.(2021)López-Úbeda, Plaza-del Arco,
  Díaz-Galiano, and Martín-Valdivia}]{lopez-ubeda_how_2021}
Pilar López-Úbeda, Flor~Miriam Plaza-del Arco, Manuel~Carlos Díaz-Galiano,
  and Maria-Teresa Martín-Valdivia. 2021.
\newblock \href {https://doi.org/10.3390/app11041838} {How {Successful} {Is}
  {Transfer} {Learning} for {Detecting} {Anorexia} on {Social} {Media}?}
\newblock \emph{Applied Sciences}, 11(4):1838.
\newblock Number: 4 Publisher: Multidisciplinary Digital Publishing Institute.

\bibitem[{MacAvaney et~al.(2021)MacAvaney, Mittu, Coppersmith, Leintz, and
  Resnik}]{macavaney_community-level_2021}
Sean MacAvaney, Anjali Mittu, Glen Coppersmith, Jeff Leintz, and Philip Resnik.
  2021.
\newblock \href {https://doi.org/10.18653/v1/2021.clpsych-1.7} {Community-level
  {Research} on {Suicidality} {Prediction} in a {Secure} {Environment}:
  {Overview} of the {CLPsych} 2021 {Shared} {Task}}.
\newblock In \emph{Proceedings of the {Seventh} {Workshop} on {Computational}
  {Linguistics} and {Clinical} {Psychology}: {Improving} {Access}}, pages
  70--80, Online. Association for Computational Linguistics.

\bibitem[{Mathur et~al.(2020)Mathur, Sawhney, Chopra, Leekha, and
  Ratn~Shah}]{mathur_utilizing_2020}
Puneet Mathur, Ramit Sawhney, Shivang Chopra, Maitree Leekha, and Rajiv
  Ratn~Shah. 2020.
\newblock \href {https://doi.org/10.1007/978-3-030-45442-5_33} {Utilizing
  {Temporal} {Psycholinguistic} {Cues} for {Suicidal} {Intent} {Estimation}}.
\newblock In \emph{Advances in {Information} {Retrieval}: 42nd {European}
  {Conference} on {IR} {Research}, {ECIR} 2020, {Lisbon}, {Portugal}, {April}
  14–17, 2020, {Proceedings}, {Part} {II}}, pages 265--271, Berlin,
  Heidelberg. Springer-Verlag.

\bibitem[{McHugh et~al.(2019)McHugh, Corderoy, Ryan, Hickie, and
  Large}]{mchugh_association_2019}
Catherine~M. McHugh, Amy Corderoy, Christopher~James Ryan, Ian~B. Hickie, and
  Matthew~Michael Large. 2019.
\newblock \href {https://doi.org/10.1192/bjo.2018.88} {Association between
  suicidal ideation and suicide: meta-analyses of odds ratios, sensitivity,
  specificity and positive predictive value}.
\newblock \emph{BJPsych Open}, 5(2):e18.

\bibitem[{Mohammadi et~al.(2019)Mohammadi, Amini, and
  Kosseim}]{mohammadi_clac_2019}
Elham Mohammadi, Hessam Amini, and Leila Kosseim. 2019.
\newblock \href {https://doi.org/10.18653/v1/W19-3004} {{CLaC} at {CLPsych}
  2019: {Fusion} of {Neural} {Features} and {Predicted} {Class} {Probabilities}
  for {Suicide} {Risk} {Assessment} {Based} on {Online} {Posts}}.
\newblock In \emph{Proceedings of the {Sixth} {Workshop} on {Computational}
  {Linguistics} and {Clinical} {Psychology}}, pages 34--38, Minneapolis,
  Minnesota. Association for Computational Linguistics.

\bibitem[{Montejo-Ráez et~al.(2024)Montejo-Ráez, Molina-González,
  Jiménez-Zafra, García-Cumbreras, and
  García-López}]{montejo-raez_survey_2024}
Arturo Montejo-Ráez, M.~Dolores Molina-González, Salud~María Jiménez-Zafra,
  Miguel~Ángel García-Cumbreras, and Luis~Joaquín García-López. 2024.
\newblock \href {https://doi.org/10.1016/j.cosrev.2024.100654} {A survey on
  detecting mental disorders with natural language processing: {Literature}
  review, trends and challenges}.
\newblock \emph{Computer Science Review}, 53:100654.

\bibitem[{Morales et~al.(2021)Morales, Dey, and Kohli}]{morales_team_2021}
Michelle Morales, Prajjalita Dey, and Kriti Kohli. 2021.
\newblock \href {https://doi.org/10.18653/v1/2021.clpsych-1.11} {Team 9: {A}
  {Comparison} of {Simple} vs. {Complex} {Models} for {Suicide} {Risk}
  {Assessment}}.
\newblock In \emph{Proceedings of the {Seventh} {Workshop} on {Computational}
  {Linguistics} and {Clinical} {Psychology}: {Improving} {Access}}, pages
  99--102, Online. Association for Computational Linguistics.

\bibitem[{Mármol~Romero et~al.(2024)Mármol~Romero, Moreno-Muñoz,
  Plaza-Del-Arco, Molina-González, and
  Montejo-Ráez}]{marmol_romero_mentalriskes_2024}
Alba~María Mármol~Romero, Adrián Moreno-Muñoz, Flor~Miriam Plaza-Del-Arco,
  M.~Dolores Molina-González, and Arturo Montejo-Ráez. 2024.
\newblock \href {https://aclanthology.org/2024.lrec-main.978/} {{MentalRiskES}:
  {A} {New} {Corpus} for {Early} {Detection} of {Mental} {Disorders} in
  {Spanish}}.
\newblock In \emph{Proceedings of the 2024 {Joint} {International} {Conference}
  on {Computational} {Linguistics}, {Language} {Resources} and {Evaluation}
  ({LREC}-{COLING} 2024)}, pages 11204--11214, Torino, Italia. ELRA and ICCL.

\bibitem[{{National Institute of Mental Health
  (NIMH)}(2025)}]{national_institute_of_mental_health_nimh_warning_2025}
{National Institute of Mental Health (NIMH)}. 2025.
\newblock \href
  {https://www.nimh.nih.gov/health/publications/warning-signs-of-suicide}
  {Warning {Signs} of {Suicide}}.

\bibitem[{Oortwijn et~al.(2021)Oortwijn, Ossenkoppele, and
  Betti}]{oortwijn_interrater_2021}
Yvette Oortwijn, Thijs Ossenkoppele, and Arianna Betti. 2021.
\newblock \href {https://aclanthology.org/2021.humeval-1.15/} {Interrater
  {Disagreement} {Resolution}: {A} {Systematic} {Procedure} to {Reach}
  {Consensus} in {Annotation} {Tasks}}.
\newblock In \emph{Proceedings of the {Workshop} on {Human} {Evaluation} of
  {NLP} {Systems} ({HumEval})}, pages 131--141, Online. Association for
  Computational Linguistics.

\bibitem[{Park et~al.(2020)Park, Park, Ahn, and Oh}]{park_suicidal_2020}
Sungjoon Park, Kiwoong Park, Jaimeen Ahn, and Alice Oh. 2020.
\newblock \href {https://doi.org/10.18653/v1/2020.emnlp-main.198} {Suicidal
  {Risk} {Detection} for {Military} {Personnel}}.
\newblock In \emph{Proceedings of the 2020 {Conference} on {Empirical}
  {Methods} in {Natural} {Language} {Processing} ({EMNLP})}, pages 2523--2531,
  Online. Association for Computational Linguistics.

\bibitem[{Patterson et~al.(1983)Patterson, Dohn, Bird, and
  Patterson}]{patterson_evaluation_1983}
William~M. Patterson, Henry~H. Dohn, Julian Bird, and Gary~A. Patterson. 1983.
\newblock \href {https://doi.org/10.1016/S0033-3182(83)73213-5} {Evaluation of
  suicidal patients: {The} {SAD} {PERSONS} scale}.
\newblock \emph{Psychosomatics}, 24(4):343--349.

\bibitem[{Paykel(1976)}]{paykel_life_1976}
Eugene~S. Paykel. 1976.
\newblock \href {https://doi.org/10.1080/0097840X.1976.9936065} {Life {Stress},
  {Depression} and {Attempted} {Suicide}}.
\newblock \emph{Journal of Human Stress}, 2(3):3--12.
\newblock Publisher: Taylor \& Francis \_eprint:
  https://doi.org/10.1080/0097840X.1976.9936065.

\bibitem[{Perrot et~al.(2022)Perrot, Hardouin, Thiabaud, Saillard,
  Grall-Bronnec, and Challet-Bouju}]{perrot_development_2022}
Bastien Perrot, Jean-Benoit Hardouin, Elsa Thiabaud, Anaïs Saillard, Marie
  Grall-Bronnec, and Gaëlle Challet-Bouju. 2022.
\newblock \href {https://doi.org/10.1556/2006.2022.00063} {Development and
  validation of a prediction model for online gambling problems based on
  players' account data}.
\newblock \emph{Journal of Behavioral Addictions}.
\newblock Section: Journal of Behavioral Addictions.

\bibitem[{Posner et~al.(2010)Posner, Brent, Lucas, Gould, Stanley, Brown,
  Fisher, Zelazny, Burke, Oquendo, and Mann}]{posner_columbia-suicide_2010}
Posner, Brent, Lucas, Gould, Stanley, Brown, Fisher, Zelazny, Burke, Oquendo,
  and Mann. 2010.
\newblock \href
  {https://cssrs.columbia.edu/wp-content/uploads/C-SSRS_Pediatric-SLC_11.14.16.pdf}
  {Columbia-{Suicide} {Severity} {Rating} {Scale} ({C}-{SSRS})}.

\bibitem[{Pourmand et~al.(2019)Pourmand, Roberson, Caggiula, Monsalve, Rahimi,
  and Torres-Llenza}]{pourmand_social_2019}
Ali Pourmand, Jeffrey Roberson, Amy Caggiula, Natalia Monsalve, Murwarit
  Rahimi, and Vanessa Torres-Llenza. 2019.
\newblock \href {https://doi.org/10.1089/tmj.2018.0203} {Social {Media} and
  {Suicide}: {A} {Review} of {Technology}-{Based} {Epidemiology} and {Risk}
  {Assessment}}.
\newblock \emph{Telemedicine and e-Health}, 25(10):880--888.
\newblock Publisher: Mary Ann Liebert, Inc., publishers.

\bibitem[{Qiu et~al.(2024)Qiu, Ma, and Lan}]{qiu_psyguard_2024}
Huachuan Qiu, Lizhi Ma, and Zhenzhong Lan. 2024.
\newblock \href {https://doi.org/10.18653/v1/2024.emnlp-main.264} {{PsyGUARD}:
  {An} {Automated} {System} for {Suicide} {Detection} and {Risk} {Assessment}
  in {Psychological} {Counseling}}.
\newblock In \emph{Proceedings of the 2024 {Conference} on {Empirical}
  {Methods} in {Natural} {Language} {Processing}}, pages 4581--4607, Miami,
  Florida, USA. Association for Computational Linguistics.

\bibitem[{Roy et~al.(2020)Roy, Nikolitch, McGinn, Jinah, Klement, and
  Kaminsky}]{roy_machine_2020}
Arunima Roy, Katerina Nikolitch, Rachel McGinn, Safiya Jinah, William Klement,
  and Zachary~A. Kaminsky. 2020.
\newblock \href {https://doi.org/10.1038/s41746-020-0287-6} {A machine learning
  approach predicts future risk to suicidal ideation from social media data}.
\newblock \emph{NPJ Digital Medicine}, 3:78.

\bibitem[{Sawhney et~al.(2021{\natexlab{a}})Sawhney, Joshi, Flek, and
  Shah}]{sawhney_phase_2021}
Ramit Sawhney, Harshit Joshi, Lucie Flek, and Rajiv~Ratn Shah.
  2021{\natexlab{a}}.
\newblock \href {https://doi.org/10.18653/v1/2021.eacl-main.205} {{PHASE}:
  {Learning} {Emotional} {Phase}-aware {Representations} for {Suicide}
  {Ideation} {Detection} on {Social} {Media}}.
\newblock In \emph{Proceedings of the 16th {Conference} of the {European}
  {Chapter} of the {Association} for {Computational} {Linguistics}: {Main}
  {Volume}}, pages 2415--2428, Online. Association for Computational
  Linguistics.

\bibitem[{Sawhney et~al.(2020)Sawhney, Joshi, Gandhi, and
  Shah}]{sawhney_time-aware_2020}
Ramit Sawhney, Harshit Joshi, Saumya Gandhi, and Rajiv~Ratn Shah. 2020.
\newblock \href {https://doi.org/10.18653/v1/2020.emnlp-main.619} {A
  {Time}-{Aware} {Transformer} {Based} {Model} for {Suicide} {Ideation}
  {Detection} on {Social} {Media}}.
\newblock In \emph{Proceedings of the 2020 {Conference} on {Empirical}
  {Methods} in {Natural} {Language} {Processing} ({EMNLP})}, pages 7685--7697,
  Online. Association for Computational Linguistics.

\bibitem[{Sawhney et~al.(2021{\natexlab{b}})Sawhney, Joshi, Gandhi, and
  Shah}]{sawhney_towards_2021}
Ramit Sawhney, Harshit Joshi, Saumya Gandhi, and Rajiv~Ratn Shah.
  2021{\natexlab{b}}.
\newblock \href {https://doi.org/10.1145/3437963.3441805} {Towards {Ordinal}
  {Suicide} {Ideation} {Detection} on {Social} {Media}}.
\newblock In \emph{Proceedings of the 14th {ACM} {International} {Conference}
  on {Web} {Search} and {Data} {Mining}}, pages 22--30, Virtual Event Israel.
  ACM.

\bibitem[{Singh~Rawat and Yu(2022)}]{singh_rawat_parameter_2022}
Bhanu~Pratap Singh~Rawat and Hong Yu. 2022.
\newblock \href {https://doi.org/10.18653/v1/2022.louhi-1.13} {Parameter
  {Efficient} {Transfer} {Learning} for {Suicide} {Attempt} and {Ideation}
  {Detection}}.
\newblock In \emph{Proceedings of the 13th {International} {Workshop} on
  {Health} {Text} {Mining} and {Information} {Analysis} ({LOUHI})}, pages
  108--115, Abu Dhabi, United Arab Emirates (Hybrid). Association for
  Computational Linguistics.

\bibitem[{Smith et~al.(2024)Smith, Peters, and Reiter}]{smith_automatic_2024}
Elke Smith, Jan Peters, and Nils Reiter. 2024.
\newblock \href {https://doi.org/10.1371/journal.pdig.0000605} {Automatic
  detection of problem-gambling signs from online texts using large language
  models}.
\newblock \emph{PLOS Digital Health}, 3(9):e0000605.
\newblock Publisher: Public Library of Science.

\bibitem[{Stene-Larsen and Reneflot(2019)}]{stene-larsen_contact_2019}
Kim Stene-Larsen and Anne Reneflot. 2019.
\newblock \href {https://doi.org/10.1177/1403494817746274} {Contact with
  primary and mental health care prior to suicide: {A} systematic review of the
  literature from 2000 to 2017}.
\newblock \emph{Scandinavian Journal of Public Health}, 47(1):9--17.
\newblock Publisher: SAGE Publications Ltd STM.

\bibitem[{Touvron et~al.(2023)Touvron, Lavril, Izacard, Martinet, Lachaux,
  Lacroix, Rozière, Goyal, Hambro, Azhar, Rodriguez, Joulin, Grave, and
  Lample}]{touvron_llama_2023}
Hugo Touvron, Thibaut Lavril, Gautier Izacard, Xavier Martinet, Marie-Anne
  Lachaux, Timothée Lacroix, Baptiste Rozière, Naman Goyal, Eric Hambro,
  Faisal Azhar, Aurelien Rodriguez, Armand Joulin, Edouard Grave, and Guillaume
  Lample. 2023.
\newblock \href {https://doi.org/10.48550/arXiv.2302.13971} {{LLaMA}: {Open}
  and {Efficient} {Foundation} {Language} {Models}}.
\newblock \emph{arXiv preprint}.
\newblock ArXiv:2302.13971 [cs].

\bibitem[{Tsui et~al.(2021)Tsui, Shi, Ruiz, Ryan, Biernesser, Iyengar, Walsh,
  and Brent}]{tsui_natural_2021}
Fuchiang~R Tsui, Lingyun Shi, Victor Ruiz, Neal~D Ryan, Candice Biernesser,
  Satish Iyengar, Colin~G Walsh, and David~A Brent. 2021.
\newblock \href {https://doi.org/10.1093/jamiaopen/ooab011} {Natural language
  processing and machine learning of electronic health records for prediction
  of first-time suicide attempts}.
\newblock \emph{JAMIA Open}, 4(1):ooab011.

\bibitem[{Turecki et~al.(2019)Turecki, Brent, Gunnell, O’Connor, Oquendo,
  Pirkis, and Stanley}]{turecki_suicide_2019}
Gustavo Turecki, David~A. Brent, David Gunnell, Rory~C. O’Connor, Maria~A.
  Oquendo, Jane Pirkis, and Barbara~H. Stanley. 2019.
\newblock \href {https://doi.org/10.1038/s41572-019-0121-0} {Suicide and
  suicide risk}.
\newblock \emph{Nature Reviews Disease Primers}, 5(1):1--22.
\newblock Publisher: Nature Publishing Group.

\bibitem[{{U.S. Food and Drug
  Administration}(2012)}]{us_food_and_drug_administration_guidance_2012}
{U.S. Food and Drug Administration}. 2012.
\newblock \href
  {https://www.fda.gov/regulatory-information/search-fda-guidance-documents/guidance-industry-suicidal-ideation-and-behavior-prospective-assessment-occurrence-clinical-trials}
  {Guidance for {Industry}: {Suicidal} {Ideation} and {Behavior}: {Prospective}
  {Assessment} of {Occurrence} in {Clinical} {Trials}}.
\newblock Publisher: FDA.

\bibitem[{Vijayakumar and Pfeifer(2020)}]{vijayakumar_self-disclosure_2020}
Nandita Vijayakumar and Jennifer~H Pfeifer. 2020.
\newblock \href {https://doi.org/10.1016/j.copsyc.2019.08.005} {Self-disclosure
  during adolescence: exploring the means, targets, and types of personal
  exchanges}.
\newblock \emph{Current Opinion in Psychology}, 31:135--140.

\bibitem[{Vries et~al.(2019)Vries, Cranenburgh, Bisazza, Caselli, Noord, and
  Nissim}]{vries_bertje_2019}
Wietse~de Vries, Andreas~van Cranenburgh, Arianna Bisazza, Tommaso Caselli,
  Gertjan~van Noord, and Malvina Nissim. 2019.
\newblock \href {https://doi.org/10.48550/arXiv.1912.09582} {{BERTje}: {A}
  {Dutch} {BERT} {Model}}.
\newblock \emph{arXiv preprint}.
\newblock ArXiv:1912.09582 [cs].

\bibitem[{Wang et~al.(2024)Wang, Zi, Zhao, Deng, and Qin}]{wang_esdm_2024}
Bichen Wang, Yuzhe Zi, Yanyan Zhao, Pengfei Deng, and Bing Qin. 2024.
\newblock \href {https://aclanthology.org/2024.lrec-main.556/} {{ESDM}: {Early}
  {Sensing} {Depression} {Model} in {Social} {Media} {Streams}}.
\newblock In \emph{Proceedings of the 2024 {Joint} {International} {Conference}
  on {Computational} {Linguistics}, {Language} {Resources} and {Evaluation}
  ({LREC}-{COLING} 2024)}, pages 6288--6298, Torino, Italia. ELRA and ICCL.

\bibitem[{Wang et~al.(2021)Wang, Fan, Shivtare, Badal, Subbalakshmi,
  Chandramouli, and Lee}]{wang_learning_2021}
Ning Wang, Luo Fan, Yuvraj Shivtare, Varsha Badal, Koduvayur Subbalakshmi,
  Rajarathnam Chandramouli, and Ellen Lee. 2021.
\newblock \href {https://doi.org/10.18653/v1/2021.clpsych-1.9} {Learning
  {Models} for {Suicide} {Prediction} from {Social} {Media} {Posts}}.
\newblock In \emph{Proceedings of the {Seventh} {Workshop} on {Computational}
  {Linguistics} and {Clinical} {Psychology}: {Improving} {Access}}, pages
  87--92, Online. Association for Computational Linguistics.

\bibitem[{Wasserman et~al.(2012)Wasserman, Rihmer, Rujescu, Sarchiapone,
  Sokolowski, Titelman, Zalsman, Zemishlany, and
  Carli}]{wasserman_european_2012}
D.~Wasserman, Z.~Rihmer, D.~Rujescu, M.~Sarchiapone, M.~Sokolowski,
  D.~Titelman, G.~Zalsman, Z.~Zemishlany, and V.~Carli. 2012.
\newblock \href {https://doi.org/10.1016/j.eurpsy.2011.06.003} {The {European}
  {Psychiatric} {Association} ({EPA}) guidance on suicide treatment and
  prevention}.
\newblock \emph{European Psychiatry}, 27(2):129--141.

\bibitem[{{World Health
  Organization}(2024)}]{world_health_organization_suicide_2024}
{World Health Organization}. 2024.
\newblock \href {https://www.who.int/news-room/fact-sheets/detail/suicide}
  {Suicide}.

\bibitem[{Zandbiglari et~al.(2025)Zandbiglari, Kumar, Bilal, Goodin, and
  Rouhizadeh}]{zandbiglari_enhancing_2025}
Kimia Zandbiglari, Shobhan Kumar, Muhammad Bilal, Amie Goodin, and Masoud
  Rouhizadeh. 2025.
\newblock \href {https://doi.org/10.1016/j.jbi.2024.104755} {Enhancing suicidal
  behavior detection in {EHRs}: {A} multi-label {NLP} framework with
  transformer models and semantic retrieval-based annotation}.
\newblock \emph{Journal of Biomedical Informatics}, 161:104755.

\bibitem[{Zhang et~al.(2024)Zhang, Zhu, Guo, Zhang, Li, and
  Hu}]{zhang_natural_2024}
Zhenwen Zhang, Jianghong Zhu, Zhihua Guo, Yu~Zhang, Zepeng Li, and Bin Hu.
  2024.
\newblock \href {https://doi.org/10.2196/58259} {Natural {Language}
  {Processing} for {Depression} {Prediction} on {Sina} {Weibo}: {Method}
  {Study} and {Analysis}}.
\newblock \emph{JMIR Mental Health}, 11:e58259--e58259.

\bibitem[{Zhong et~al.(2019)Zhong, Mittal, Nathan, Brown, Knudson~González,
  Cai, Finan, Gelaye, Avillach, Smoller, Karlson, Cai, and
  Williams}]{zhong_use_2019}
Qiu-Yue Zhong, Leena~P. Mittal, Margo~D. Nathan, Kara~M. Brown, Deborah
  Knudson~González, Tianrun Cai, Sean Finan, Bizu Gelaye, Paul Avillach,
  Jordan~W. Smoller, Elizabeth~W. Karlson, Tianxi Cai, and Michelle~A.
  Williams. 2019.
\newblock \href {https://doi.org/10.1007/s10654-018-0470-0} {Use of natural
  language processing in electronic medical records to identify pregnant women
  with suicidal behavior: towards a solution to the complex classification
  problem}.
\newblock \emph{European Journal of Epidemiology}, 34(2):153--162.

\bibitem[{Zirikly et~al.(2019)Zirikly, Resnik, Uzuner, and
  Hollingshead}]{zirikly_clpsych_2019}
Ayah Zirikly, Philip Resnik, Özlem Uzuner, and Kristy Hollingshead. 2019.
\newblock \href {https://doi.org/10.18653/v1/W19-3003} {{CLPsych} 2019 {Shared}
  {Task}: {Predicting} the {Degree} of {Suicide} {Risk} in {Reddit} {Posts}}.
\newblock In \emph{Proceedings of the {Sixth} {Workshop} on {Computational}
  {Linguistics} and {Clinical} {Psychology}}, pages 24--33, Minneapolis,
  Minnesota. Association for Computational Linguistics.

\bibitem[{Zogan et~al.(2022)Zogan, Razzak, Wang, Jameel, and
  Xu}]{zogan_explainable_2022}
Hamad Zogan, Imran Razzak, Xianzhi Wang, Shoaib Jameel, and Guandong Xu. 2022.
\newblock \href {https://doi.org/10.1007/s11280-021-00992-2} {Explainable
  depression detection with multi-aspect features using a hybrid deep learning
  model on social media}.
\newblock \emph{World Wide Web}, 25(1):281--304.
\newblock Company: Springer Distributor: Springer Institution: Springer Label:
  Springer Number: 1 Publisher: Springer US.

\end{thebibliography}
\end{document}